# MOANA: Multi-Objective Ant Nesting Algorithm for Optimization Problems

**Noor A. Rashed [1]*, Yossra H. Ali [2] Tarik A. Rashid [3,] * and Seyedali Mirjalili [4]**

[1] Computer Sciences Dept., Univ. of Technology, Baghdad, Iraq; cs.20.62@grad.uotechnology.edu.iq
[2] Computer Sciences Dept., Univ. of Technology, Baghdad, Iraq; Yossra.h.ali@uotechnology.edu.iq
[3] Computer Sciences & Engineering Dept., Artificial Intelligence Centre and Innovation, Univ. of Kurdistan Hewler, Iraq; tarik.ahmed@ukh.edu.krd
[4] Centre for Artificial Intelligence Research and Optimisation, Torrens University,
Brisbane, QLD 4006, Austral, QLD 4006, Austral. University Research and Innovation Center (EKIK), Obuda University, Budapest, 1034, Hungary; ali.Mirjalili@torrens.edu.au
* Correspondence: cs.20.62@grad.uotechnology.edu.iq (N.A.R.); tarik.ahmed@ukh.edu.krd (T.A.R.);

**Abstract:** This paper presents the Multi-Objective Ant Nesting Algorithm (MOANA), a novel extension of the Ant Nesting Algorithm (ANA), specifically designed to address multi-objective optimization problems (MOPs). MOANA incorporates adaptive mechanisms, such as deposition weight parameters, to balance exploration and exploitation, while a polynomial mutation strategy ensures diverse and high-quality solutions. The algorithm is evaluated on standard benchmark datasets, including ZDT functions and the IEEE Congress on Evolutionary Computation (CEC) 2019 multi-modal benchmarks. Comparative analysis against state-of-the-art algorithms like MOPSO, MOFDO, MODA, and NSGA-III demonstrates MOANA's superior performance in terms of convergence speed and Pareto front coverage. Furthermore, MOANA's applicability to real-world engineering optimization, such as welded beam design, showcases its ability to generate a broad range of optimal solutions, making it a practical tool for decision-makers. MOANA addresses key limitations of traditional evolutionary algorithms by improving scalability and diversity in multi-objective scenarios, positioning it as a robust solution for complex optimization tasks.

## 1. Introduction

In many real-world scenarios, optimization problems involve multiple conflicting objectives that must be addressed simultaneously. Such problems are known as multi-objective optimization problems (MOPs) and arise in various fields, including engineering, environmental science, logistics, and economics. [1]. The complexity of these problems lies in the fact that optimizing one objective often compromises another, necessitating careful trade-offs. The goal is not to find a single solution but rather a set of optimal solutions known as the Pareto front, where no objective can be improved without negatively impacting another [2].

Over the years, multi-objective evolutionary algorithms (MOEAs) have become the standard approach for addressing MOPs due to their ability to approximate the Pareto front while maintaining diversity among solutions. However, despite their success, traditional MOEAs such as Non-Dominated Sorting Genetic Algorithm II (NSGA-II), Multi-Objective Particle Swarm Optimization (MOPSO)[3, 4], and Differential Evolution (DE) faces significant challenges[5]. These challenges include slow convergence, poor scalability in high-dimensional problems, and difficulty in maintaining diversity across the Pareto front.

In this study, ANA was rationally selected over other algorithms, such as DE and CMA-ES, due to its adaptive mechanisms, which allow for a more balanced exploration-exploitation trade-off. Unlike DE or CMA-ES, which depend on mutation and recombination to navigate the solution space, ANA employs a deposition weight mechanism that dynamically adjusts the agent's behavior. This enables MOANA to achieve better coverage of the Pareto front and discover both global and local optima more efficiently, which is critical in engineering optimization problems like welded beam design.

To address these limitations, this paper introduces the Multi-Objective Ant Nesting Algorithm (MOANA), a novel extension of the Ant Nesting Algorithm (ANA)[6]. MOANA is designed to balance exploration and exploitation dynamically, ensuring better coverage of the Pareto front while enhancing convergence speed. By integrating mechanisms such as adaptive deposition weight parameters and polynomial mutation strategies, MOANA offers a scalable and efficient solution to multi-objective problems. Additionally, the practical application of MOANA in real-world engineering problems, such as the design of welded beams, highlights its effectiveness and versatility.

The main contributions of this work include:

1) The development of MOANA as a multi-objective extension of the single-objective ANA algorithm.
1. The balance of exploration and exploitation in MOANA, achieved through the deposition weight parameter, improves both coverage and convergence speed.
2. The use of an archive technique for non-dominant solutions ensures that high-quality solutions are stored effectively.
3. A polynomial mutation strategy that guarantees diversity in solutions, with stored deposition weights (dw) for reuse in future iterations.
4. The enhancement of algorithm performance through the integration of Hypercube grids, which provide a mechanism to select local and global guide individuals.
5. The successful application of MOANA to practical engineering problems, such as the optimization of welded beam designs, demonstrates its versatility.

This paper evaluates MOANA on standard benchmark datasets, including the ZDT functions and the IEEE Congress on Evolutionary Computation (CEC) 2019 multi-modal benchmarks. Comparisons against state-of-the-art MOEAs show MOANA's superior performance in terms of convergence speed and coverage of the Pareto front. Additionally, the practical application of MOANA in real-world engineering problems, such as the design of welded beams, highlights its effectiveness and versatility.

The organization of this paper is as follows: Section 2 begins with a comprehensive literature review, exploring existing multi-objective optimization algorithms such as NSGA-II, MOPSO, DE, and CMA-ES, along with their limitations. This section sets the context for the introduction of MOANA, highlighting the specific challenges these traditional algorithms face, including slow convergence and difficulty in maintaining diversity across the Pareto front. Following the review, the theoretical framework is detailed, including the definitions of Pareto sets and Pareto fronts. The section concludes with a mathematical description of the Multi-Objective Ant Nesting Algorithm (MOANA), showcasing its novel features like adaptive exploration-exploitation balancing and deposition weight mechanisms.
In Section 3, the results of applying MOANA to various benchmark problems are presented and analyzed, with a comparison to state-of-the-art algorithms such as NSGA-II, MOPSO, and MODA. The discussion highlights the strengths of MOANA in terms of convergence speed, solution diversity, and Pareto front coverage. Section 4 demonstrates the practical application of MOANA by solving a real-world engineering challenge: the optimization of a welded beam design. This section emphasizes how MOANA’s capabilities translate into effective decision-making for complex engineering problems. Finally, Section 5 presents the concluding remarks, summarizing the main findings and suggesting directions for future work, including the potential for MOANA to address more diverse and high-dimensional optimization problems in various domains.

## 2. Literature Review

The literature on multi-objective evolutionary algorithms (MOEAs) covers a wide range of approaches for solving multi-objective optimization problems (MOPs). These algorithms are typically categorized into indicator-based evolutionary algorithms (IBEAs)[7], decomposition-based algorithms[8], and dominance-based algorithms[9]. Each of these approaches addresses MOPs by estimating Pareto-optimal solutions. For example, dominance-based algorithms such as NSGA-II [10]and SPEA2[11] sort populations using non-dominated sorting techniques, while decomposition-based algorithms[12] break down objectives by applying weights to approximate the Pareto front. Among these, NSGA-II is widely recognized for its ability to maintain diversity through crowding distance, and it has been extended to tackle problems with more complex objectives, such as in NSGA-III [13] and MOPSO[4], another popular approach, uses an archive grid to ensure diversity while maintaining computational efficiency.

In the context of population-based metaheuristics[14], Differential Evolution (DE) and Covariance Matrix Adaptation In the context of population-based metaheuristics, Differential Evolution (DE) and Covariance Matrix Adaptation Evolution Strategy (CMA-ES) are well-regarded for real-parameter, single-objective optimization. While DE maintains diversity, CMA-ES struggles with stagnation at local optima[15]. To address this, a hybrid algorithm, IR-CMA-ES, combines CMA-ES with DE to enhance exploration and prevent stagnation. This hybrid approach improves performance across benchmarks, particularly in maintaining diversity. However, IR-CMA-ES has limitations, including increased computational cost and sensitivity to parameter settings. It may also face scalability issues and slower convergence in highly complex problems, as it may not always balance exploration and exploitation optimally[16]. Over the past decade, several other MOEAs have emerged, such as the Whale Optimization Algorithm[17], Ant Lion Optimizer[18], Grey Wolf Optimizer[19], Moth Flame Optimization[20], Multi-Objective Cat Swarm Optimization[21], Dragonfly algorithm[22], a multi-objective learner performance-based behavior algorithm[23], and multi-objective fitness-dependent optimizer[24]. These methods have introduced novel mechanisms inspired by biological or physical processes to address the limitations of earlier MOEAs. For example, IBEAs have gained popularity due to their strong theoretical foundations, automatically addressing convergence and diversity issues through the use of performance indicators like hypervolume and epsilon-indicator[25, 26]. However, despite the progress, MOPs still require algorithms that can balance two critical factors: the precision of the Pareto-optimal solutions and the diversity of the anticipated solutions. Most approaches rely on meta-heuristic methods, which start with random solutions and gradually improve them over time[27]. The challenge, however, remains in balancing exploration (searching for new solutions) and exploitation (refining existing ones), and several researchers have proposed techniques to address this.

For instance, Mostaghim and Teich[28] introduced a sigma approach for selecting local guides, while Pulido and Coello [29] used clustering to enhance diversity in the Pareto front. Other researchers, such as Zitzler[30], focused on the importance of elitism by using crossover and mutation from both population and repository individuals[27], while Laumanns et al[31]. applied e-box dominance to improve convergence and diversity simultaneously. Several hybrid methods, such as those combining Grey Wolf Optimization for industrial systems or altered genetic algorithms for ship scheduling and routing[32, 33], have also shown success in specific real-world challenges[34]. Additionally, algorithms like Pareto Entropy MOPSO, which selects global guides based on individual density, have been applied to complex industrial designs, further advancing the field[35]. Several design issues, including those involving a pressure vessel, speed reducer,four-bar truss, coil compression spring, and automobile side collision, were addressed using the new MOLPB technique[23].

Five categories of understanding make up cultural algorithms, according to[24]: Historical knowledge, topographical, standard, domain, and Situational knowledge. These file types are explained briefly in the list below:

1) historical knowledge: records crucial events in the investigation by tracing the historical background of important individuals. Key events can encompass substantial alterations in the search environment or a notable transformation in the search domain. Individuals utilize historical data to select a favored course of action.
2) Topographical knowledge: divides easily accessible search landscapes into cells. Each cell represents a distinct spatial characteristic and selects the best individual within its range. Simply said, geographical knowledge guides persons to the optimal cell.
3) standard knowledge presents promising decision variable ranges. Individual adjusting strategies are provided. It draws people into a decent range.
4) Domain knowledge refers to knowledge that is recorded and utilized to assist in the search process related to a specific problem domain.
5) Situational knowledge refers to a collection of information and insights that help in understanding and making sense of the experiences of a certain group of individuals. Situational knowledge drives individuals towards exemplars, which can be either local or global authorities in their respective fields.

In response to the limitations of traditional MOEAs, the Multi-Objective Ant Nesting Algorithm (MOANA) was developed as a novel approach to balance exploration and exploitation more adaptively. MOANA builds upon the Ant Nesting Algorithm (ANA) by incorporating deposition weight parameters to guide the search process and polynomial mutation strategies to maintain diversity in the population. Compared to other algorithms like NSGA-II, MOPSO, and MOFDO, MOANA offers faster convergence, better Pareto front coverage, and improved scalability, especially in high-dimensional optimization tasks. Moreover, its application in real-world scenarios, such as welded beam design, underscores its practical utility in generating a broad range of optimal solutions for decision-makers. MOANA's innovative approach to multi-objective optimization addresses key challenges faced by traditional algorithms, making it a valuable tool in tackling complex, multi-faceted problems across various domains.

## 3. Methodology

The following section provides an overview of the preliminary concepts and fundamental definitions related to multi-objective optimization. MOANA is comprehensively described in terms of mathematical and programmatic aspects.

### *3.1 Pareto optimal solutions set*

Several real-world challenges involve multiple conflicting objectives. These types of problems require the use of optimization with multiple objectives. Due to conflicting goals, a solution that is severe in one aspect must be balanced in another aspect. An illustrative instance of a predicament with several objectives is the task of acquiring an automobile. The resolution of a two-objective problem entails achieving an equilibrium between the factors of cost and comfort. The objective of optimization is to identify the optimal solution that achieves the most favorable trade-off among all the objectives[36]. From a mathematical perspective, MOPs can be accurately expressed in the following manner, without any reduction in ubiquity as see in equation (1).

$$\text{Minimize}: F(\vec{x}) = \{f_1(\vec{x}), f_2(\vec{x}), \dots, f_n(\vec{x})\}$$
$$\text{subject to:}$$
$$g_{i(\vec{x})} \le 0, i = 1,2,\dots,m \quad (1)$$
$$h_{i(\vec{x})} \le 0, i = 1,2,\dots,\text{p}$$

h and g are the rules, m represents an inequality rule, p represents an equality rule, and n represents the number of objectives[37]. (MOPs) entail attempting to maximize the number of conflicting goals. Since there might not be a single solution that simultaneously optimizes all goals, solving MOPs could be challenging. Rather, one way to solve this problem is to use a set of trade-off solutions referred to as the Pareto front. Representing the optimal balance between goals. To improve readability, this subsection will give a brief introduction to Pareto optimality. Pareto optimality is the term for solutions that may be explained by applying the following definitions[38].

- Def. #1: for vectors (solution) $\vec{a}$ & $\vec{b}$ in optimization problem $K^t$.
- For $i = 1, 2, \dots, m, \vec{a} \le \vec{b}$ if the goals of the vector $a^a$ equal to or smaller than the objectives of the vector $\vec{b}$ and at least there is $\vec{a_i} < \vec{b_i}$.
- Def. #2: if $\vec{a} \le \vec{b}$ then: $\vec{a}$ dominates $\vec{b}$, and indicated by $\vec{a} \prec \vec{b}$.
- Def. #3: 2 solutions could not dominate one another if Def. #1 isn't applied, in such case, solutions $\vec{a}$ and $\vec{b}$ are non-dominated concerning one another, and indicated as $\vec{a} \nprec \vec{b}$ is the set of all c solution set $P_s$, and $P_s := \{\vec{a}, b \in A \mid \exists F(a) \succ F(b)\}$ .
- Def #4: The set that holds equivalent object values of the Pareto optimal solutions in Ps, is referred to as the Pareto optimal front $P_f$, and $P_f := \{F(\vec{a}) | \vec{a} \in |P_s\}$.

### *3.2 Multi-objective Ant Nesting Algorithm*

This algorithm is derived from a single-objective Ant Nesting Algorithm[6]. The ANA algorithm is a Leptothorax-inspired swarm intelligent algorithm. It optimizes real-world issues by replicating ant nesting. Each search agent represents an ant and explores and exploits the search space. Agents use Pythagorean theorem-based rules to determine movement direction and distance. The artificial worker ant in the ANA algorithm deposition position is updated using equation (2), which is also used in MOANA[6].

$$X_{t+1,i} = X_{t,i} + \Delta X_{t+1,i} \quad (2)$$

In which ($X_{t,i}$) represent the deposition position of an artificial worker ant, ($t$) represents current iteration, t= {1,2,3,4…n}, ($i$) represent current worker ant, i= {1,2,3,4…m} and ($\Delta X_{t+1,i}$) represent the change rate. The difference between the deposition position of the current worker ant ($X_{t,i}$) and the position of deposition of the local best-known worker ant, ($X_{t,ibest}$) determines the rate of change of deposition position ($\Delta X_{t+1,i}$) simulating leaning toward the most dropped building material. To improve its deposition solution, each worker ant goes toward the best-known worker ant. The following equations (3), (4), and (5) calculate the ($\Delta X_{t+1,i}$):

$$\Delta X_{t+1,i} = dw \times (X_{t,ibest} - X_{t,i}) \quad (3)$$

The next rule is followed for the calculation of the value of ($\Delta X_{t+1,i}$) when the current worker ant is the best-known local ant.

$$\Delta X_{t+1,i} = r \times X_{t,i} \quad (4)$$

the current deposition position equals the previous one.

$$\Delta X_{t+1,i} = r \times (X_{t,ibest} - X_{t,i}) \quad (5)$$

The ANA algorithm's deposition weight ($dw$) is a mathematical description of a worker ant's random walk and depends on its prior ($T_{\text{previous}}$ ) and current ($T$) tendency rate to deposit grain at some certain spot. The slope sides of the difference between worker ants' current and previous deposition positions to best-found position in the Pythagorean theorem are T and T previous, with their fitness difference being the other sides as in equations (6), (7), and (8).

$$dw = r \times \left(\frac{T}{T_{\text{previous}}}\right) \quad (6)$$

where, $r$ represent a random number in a range of $[-1,1]$, works as a factor of deposition to control $dw$. worker ant's tendency rate of deposition ($T$) is calculated as follows:

$$T = \sqrt{\left(X_{t,ibest} - X_{t,i}\right)^2 - \left(X_{t,ibest}\text{ fitness } - X_{t,i}\text{ fitness }\right)^2} \quad (7)$$

The worker ant's previous tendency deposition rate $\left(T_{\text{previous}}\right)$ can be estimated as follows:

$$T_{previous} = \sqrt{\left(X_{t,\ ibest} - X_{t,\ iprevious}\right)^2 - \left(X_{t,\ ibest}\,fitness - X_{t,\ iprevious}\ fitness\right)^2} \quad (8)$$

For more detailed information about single-objective ANA, and other single-objective interested readers are directed to references[6]. The algorithmic structure of MOANA has some resemblance to that of a single objective ANA, it incorporates some supplementary enhancements as follows:

1) An archive, often referred to as a repository, is commonly employed in optimization to save Pareto front solutions. This practice has been well-documented in the existing literature[39].
2) The MOANA method employs equation (9) and (10).

$$T = \sum_{o=1}^{n} \sqrt{\left(X_{t,ibest} - X_{t,i}\right)^2 - \left(X_{t,ibest}\text{ fitness } - X_{t,i}\text{ fitness }\right)^2} \quad (9)$$

$$T_{\text{previous=}} \sum_{o=1}^{n} \sqrt{\left(X_{t,ibest} - X_{t,i\text{previous}}\right)^2 - \left(X_{t,ibest}\text{ fitness } - X_{t,i\text{previous}}\text{ fitness }\right)^2} \quad (10)$$

In lieu of equation (9-10), where the

$$\textstyle\sum_{o=1}^{n} \sqrt{\left(X_{t,ibest} - X_{t,i}\right)^2 - \left(X_{t,ibest}\text{ fitness } - X_{t,i}\text{ fitness }\right)^2}$$

represent sum of global best individuals where n is denoted by number of objectives, o= [1,2,3, 4, ……n], and the

$$\textstyle\sum_{o=1}^{n} \sqrt{\left(X_{t,ibest} - X_{t,i\text{previous}}\right)^2 - \left(X_{t,ibest}\text{ fitness } - X_{t,i\text{previous}}\text{ fitness }\right)^2}$$

represent the sum of previous individuals' global best individual again, n denoted by the number of objectives, o= [1,2,3, 4……. n], due to its ability to compute all generated solutions and yield suitable solutions. The Table 1 represent all notation and definitions of key concepts of MOANA algorithm.

**Table 1.** Notation and Definitions of Key Concepts of MOANA

| Notation | Definitions |
|---|---|
| $i$ | current worker ant |
| $t$ | current iteration |
| $m$ | No. of iterations |
| $n$ | No. of an artificial worker ant |
| $X_{t,i}$ | The current deposition position of worker ant |
| $X_{t,ibest}$ | The deposition position of the local best worker ant |
| $X_{t,\text{ iprevious}}$ | The previous deposition position of worker ant |
| $X_{t,i}$ fitness | The fitness of the current deposition position of worker ant |
| $X_{t,ibest}$fitness | The fitness of the best local deposition position of worker ant |
| $X_{t,\text{ iprevious}}$ fitness | The fitness of the previous deposition position of worker ant |
| $\sum_{n=1}^{o} X_{t,i}$ fitness | The summation of the fitness of the current deposition position of worker ant |
| $\sum_{p=1}^{o} X_{t,ibest}$ fitness | The summation of the fitness of the best local deposition position of a worker ant |
| $\sum_{n=1}^{o} X_{t,\text{ iprevious}}$ fitness | The summation of the fitness of the previous deposition position of a worker ant |
| $T$ | The tendency rate of the current deposition position of a worker ant |
| $T_{\text{previous}}$ | The tendency rate of the previous deposition position of a worker ant |
| $\Delta X_{t+1,i}$ | the change rate of deposition position of a worker ant |
| $X_{t+1,i}$ | the new position of a worker ant |
| $r$ | a random number in a range of $[-1,1]$ |
| $dw$ | The deposition weight |
| $\beta_{\max}(x_j)$ | the largest allowable perturbation between the modified and original solution |
| $S_{Ni}$ | new solution |
| $S_i(x_j)$ | present solution |
| $NP$ | population size |
| $q$ | a positive real integer |
| $v$ | a random number uniformly distributed between 0 and 1 |
| $l$ | the boundary of decision variable $x$ |
| $u$ | upper boundary of decision variable $x$ |
| $n$ | the number of decision variables (problem dimensions) |

In the context of MOPs, it is not feasible to select the fittest solution as a universal guide, as is commonly done in single-objective optimization. This is due to the presence of many objectives in MOPs. Typically, these aims compete with one another. Hence, the process of choosing a worldwide guide necessitates a more discerning and deliberate approach. To achieve this objective:

1) A mechanism known as the controller for archives is employed to partition the repository into numerous squares of similar sizes.
2) Prior to being included to the archive as a non-dominated solution, Pareto front solutions undergo a polynomial mutation to ensure the diversity in the solutions.
3) In MOEAs[40], the polynomial mutation was used as variation operator; Additional storage has been allocated to store the previous dw for future reuse in subsequent iterations, hence enhancing the performance of the process. Its definition found in [41]Equation (11).

$$
\begin{aligned}
&S_i = (x_1, x_2, \dots, x_n) \\
&S_{Ni}(x_j) = S_i(x_j) + \alpha \cdot \beta_{max}(x_j), i = 1,2,\dots,\mathrm{NP}, \\
&j = 1,2,\dots n \\
&\alpha = \begin{cases} (2v)^{\frac{1}{(q+1)}} - 1, v < 0.5 \\ 1 - (2(1-v))^{\frac{1}{(q+1)}}, \text{otherwise} \end{cases} \\
&\beta_{max}(x_j) = \mathrm{Max}\left[S_i(x_j) - l_j, u_j - S_i(x_j)\right] \\
&i = 1,2,\dots,\mathrm{NP}, j = 1,2,\dots n
\end{aligned}
\tag{11}
$$

Where $\beta_{\max}(x_j)$ represent the largest allowable perturbation between the modified and original solution, $S_{Ni}$ represent new solutions, $S_i(x_j)$ represent the present solution, $NP$ denotes the population size. $q$ denotes a positive real integer. $v$ represents a random number uniformly distributed between 0 and 1. $l$ denotes the lower boundary of the decision variable $x$. $u$ denotes the upper boundary of the decision variable $x$. $n$ denotes the number of decision variables (problem dimensions) as seen in the algorithm (1).

| Algorithm (1): Polynomial Mutation Operator |
|---|
| 1. Initialize<br>let $S_i = (x_1, x_2, \dots, x_n)$ be the current solution, where n is the number of decision variables.<br>$S_{Ni}(x_j)$ is the mutated solution, and $l_j$ and $u_j$ are the lower and upper boundary for the decision variable $x_j$,respectively<br>Define q(distribution index for mutation) and a random number v uniformly distributed between 0 and 1.<br>2. Calculate Maximum Perturbation<br>Compute $\beta_{max}(x_j) = \mathrm{Max}\left[S_i(x_j) - l_j, u_j - S_i(x_j)\right]$<br>3. Determine Mutation Step<br>For each decision variable $x_j$, calculate the mutation step based on the value of v<br>$\alpha = \begin{cases} (2v)^{\frac{1}{(q+1)}} - 1, v < 0.5 \\ 1 - (2(1-v))^{\frac{1}{(q+1)}}, \text{otherwise} \end{cases}$<br>4. Apply mutation<br>Update the value of $S_{Ni}(x_j)$<br>$S_{Ni}(x_j) = S_i(x_j) + \alpha \cdot \beta_{max}(x_j), i = 1,2,\dots,\mathrm{NP},$ |

5. Return Mutated Solution

Unlike single-objective optimization, where the fittest solution can be utilized as a global guide or standard knowledge, multi-objective optimization problems have several objectives typically, these objectives conflict. Therefore, selecting a global guide necessitates further consideration. Within the framework of this investigation, the worldwide manual The individual is the optimal solution chosen from the least populated area using an artificial ant worker, similar to the approach used in MOPSO[42]. In multi-dimensional problems, the archive controller partitions the archive into grids of similar size, called sub-hyperspheres. The grids used in the MOANA system are known as hypercube grids and they symbolize the topographical utilization of knowledge. The method can identify the least populated area by counting the number of solutions within every grid thanks to the hypercube grid approach[43]. The area with the lowest population density will provide the best option, as depicted in Figure 1.

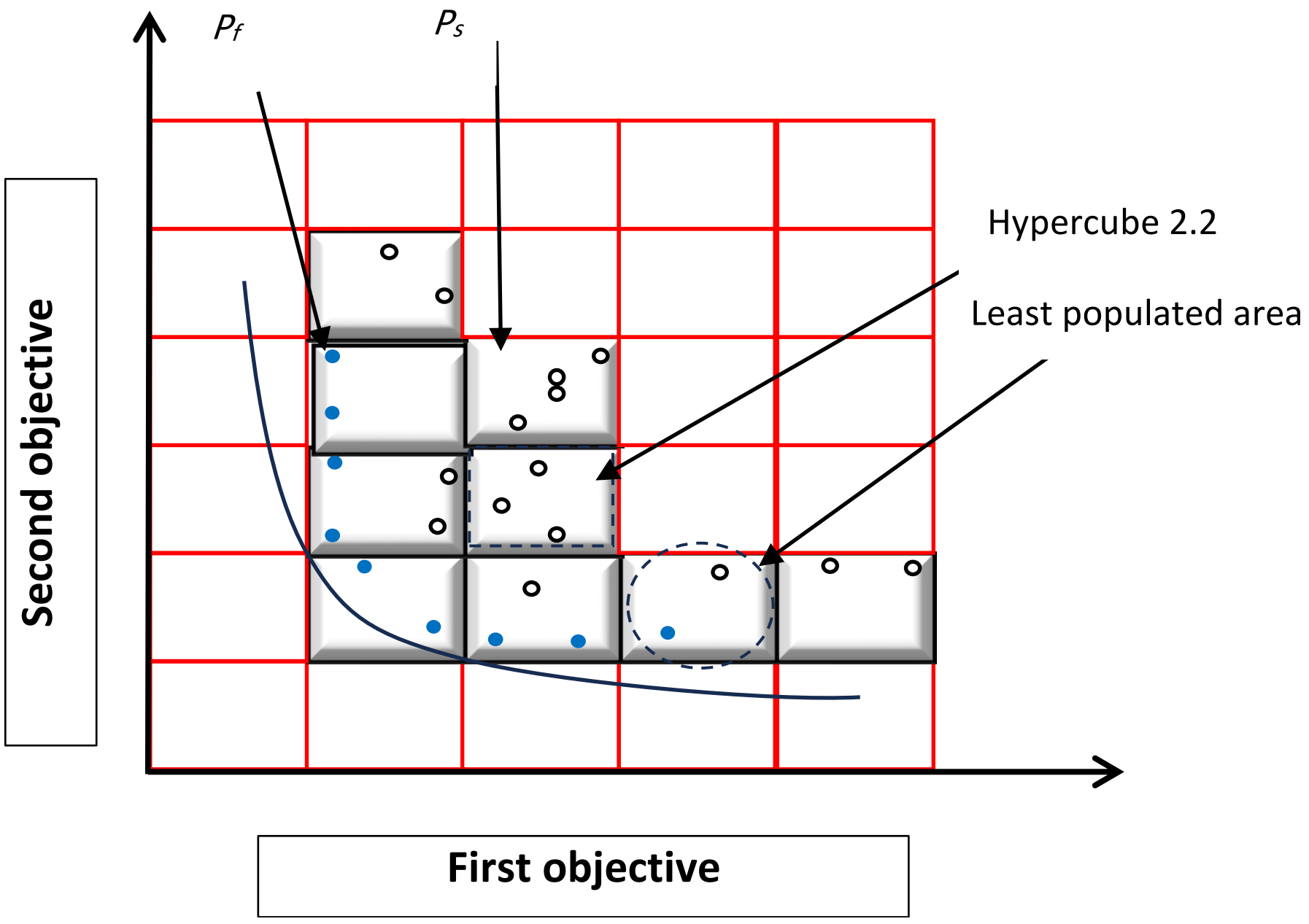

**Figure 1.** Pareto solution, Pareto front, and hyper-cube grids aid in picking global and local guides

Figure 1. Represents the Pareto Front($PF$) and the Pareto Set ($PS$)in a multi-objective optimization scenario, which is essential for identifying trade-offs between conflicting objectives. Each point in the figure corresponds to a potential solution to the optimization problem, and the grid structure (Hypercube) divides the search space, helping guide the algorithm toward unexplored regions.

The least populated area within the grid highlights a zone where fewer solutions have been identified, indicating areas of the search space that may still hold unexplored, potentially optimal solutions. MOANA aims to balance exploration and exploitation within these regions, ensuring a comprehensive search of the Pareto front while maintaining diversity among solutions. To sustain diversity in Pareto front solutions, the algorithm selects a global guide from the least populated area. This approach provides decision-makers with a wider range of options. However, due to the archive's size constraints, the archive controller removes the solution with the lowest quality from the grid that contains the greatest number of solutions when a new solution that is not dominated by others is found and the archive is already full. If the new solution outperforms the weakest solution in the archive, it will be accommodated. To select individual guides based on situational knowledge, a hypercube grid is used to partition the search space into cells of equal size. Within each cell, the best individual solution is chosen as a local reference.

### *3.3 Mechanism of Multi-objective Ant Nesting Algorithm*

The proposed approach integrates an Ant nesting algorithm, The MOANA algorithm initiates by stochastically dispersing search agents throughout the designated search area. In the given algorithm (2) a hybrid grid, and polynomial mutation techniques to identify non-dominated solutions, also known as the Pareto front.

| Algorithm (2): MOANA |
|---|
| **1-** Initialize worker ant population randomly $X_i$ (i=1,2,3 …. N).<br>**2-** Initialize worker ant previous position $X_i$ previous.<br>**3-** Creating an archive for non-dominated solutions with specific sizes.<br>**4-** Generate Hybrid cube Grid<br>**5-** While (t) iteration limit not reached (m)<br>or solution good enough.<br>**6-** For every one of the artificial workers ant $X_{t,i}$<br>**7-** Find the best artificial worker ant $X_{t,best}$<br>**8-** Generate random walk r in [-1,1] range<br>**9-** If ($X_{t,i} == X_{t,ibest}$)<br>**10-** Estimate $\Delta \boldsymbol{X}_{t+1,i}$ Using Eq (4)<br>**11-** Else if ($X_{t,i} = X_{t,i}$ previous)<br>**12-** Calculate $\Delta \boldsymbol{X}_{t+1,i}$ using Eq (5)<br>**13-** Else<br>**14-** Calculate $\boldsymbol{T}$ using Eq (9)<br>**15-** Estimate T previous using Eq (10)<br>**16-** Estimate dw using Eq (6) → minimization problem<br>**17-** Calculate $\Delta \boldsymbol{X}_{t+1,\,i}$ using Eq (3)<br>**18-** End if<br>**19-** Calculate $\Delta \boldsymbol{X}_{t+1,\,i}$ Using Eq (2)<br>**20-** If ($X_{i,t+1}$ fitness dominate on $X_{i,t}$ fitnesses)<br>**21-** Move accepted and $X_{ti,t}$ assigned to $X_{i,t}$ previous, and saved dw<br>**22-** Else<br>Maintain current position// (do not move)<br>**23-** EndIf<br>**24-** Apply polynomial mutation<br>**25-** Add non–dominated ants (solutions) to the archive.<br>**26-** Keep only non–dominated members in the archive<br>**27-** Update Hypercube Grid indices<br>End for<br>End while |

At the outset, a group of worker ants is created, every one of which represents one of the potential solutions to the optimization problem at the line (1). Create the previous locations of these insects in line (2). To keep note of unbeatable answers, a meticulous archive is maintained. This archive represents the non-dominate solutions with specific sizes in line (3). A hybrid cube grid is proposed as a method to divide the search space into distinct cells or locations, which would facilitate structured research in line 4. In this algorithm, the main loop starts at line (5) until iterations are not reached or some condition is stopped. Using an iterative approach, the program handles each artificial worker ant individually. The operation begins in lines (6-27), identifies the best ant in the current population, makes random adjustments to facilitate discovery, and employs problem-specific equations to determine positional changes. When an ant discovers a new location that is more suitable for its requirements than its previous location, it notifies its previous location, and updates it, while also keeping track of its position change. If the ant's fitness level does not improve in response to the new circumstances, it will remain in its previous position in equations (2), (3), (4), (5), and (6) in line (9), to find the best individuals for search agent from the line (10-16), apply conditions to find the global best worker ant using equations (2,3,4, and 6), then calculate the tendency and previous tendency for dw using equations (9), and (10) to compute a new search agent using equation (2) at line (19). If the result of the new best worker ant fitness is dominated by the current result of the worker ant fitness (cost function), accept the new result and save the dw for potential reuse for the next iterations. If it is not found, use the previous dw for the previous position instead of the new result and maintain the current result for the search agent in the hope to find better results in the next iterations this is found in lines (20 -23). A polynomial mutation technique is used to diversify the search procedure in line (25). Ants whose locations are non-dominated are kept in the archive, as are all non-dominated solutions in line (26). Following the preceding procedure, the values of the hypercube grid indices are updated in line (27). Its primary objective is to discover optimal solutions to problems involving multi-objective optimization. More explanations are given in Figure 2. In the section (Results and Discussion).

### *3.4 Multi-objective Ant Nesting Algorithm Time and Space Complexity*

The computational complexity of MOANA, in both time and space, reflects its scalability and efficiency in handling complex multi-objective optimization tasks. MOANA's time complexity per iteration depends on the population size (n), the dimensionality of the problem (d), the cost of evaluating the objective function (CF), and additional operations such as deposition weights (dw) and polynomial mutation (pm), which enhance exploration and exploitation. For each iteration, the time complexity can be expressed as O (n * d + n * CF + n * dw + n * pm), meaning the total time complexity is proportional to the number of iterations (T), resulting in O (T* (n * d + n * CF + n * dw + n * pm)). This ensures that MOANA's time complexity grows linearly with the population size and the number of iterations, making it more scalable compared to algorithms like NSGA-III, which can have quadratic complexity of $O(n^2)$ in some cases.

Regarding space complexity, MOANA efficiently manages memory usage by storing only the current population and the deposition weights that are reused in the next iteration. The space complexity can be represented as O(n*d + dw*d), where n*d accounts for the decision variables in the current population, and dw*d represents the space required for storing the deposition weights. Since the deposition weights are relatively small compared to the population size, the space complexity remains manageable throughout the iterations. This allows MOANA to maintain an efficient and scalable memory profile, ensuring that the algorithm can handle large-scale optimization problems without significant memory overhead. Compared to other algorithms like MOFDO, which also has linear time complexity (O(p*n + p*CF)), MOANA's complexity is slightly higher due to the additional operations for deposition weights and polynomial mutation. However, it remains less complex than algorithms like MOPSO and MODA, which involve more computational overhead due to the need to calculate global and local bests or various weight parameters.

MOFDO, for example, has a simpler calculation mechanism, requiring only the calculation of a random number and fitness weight for each agent. In contrast, MOPSO requires additional parameters like global best, agent best, and random search factors (C1, C2, R1, R2). MODA involves even more complex calculations, such as attraction, distraction, alignment, cohesion, and separation weights, which depend on the values of other agents, leading to cumulative computations. NSGA-III, with its $O(ng * no * np^2)$ complexity, is more computationally demanding than MOFDO, MOANA, and other linear complexity algorithms. This makes MOANA's linear complexity in both time and space a highly scalable algorithm, capable of efficiently solving large-scale, real-world optimization problems without incurring excessive computational costs.

## 4. The Results and Discussion

Two discrete classes have been chosen for multi-objective test functions to assess the effectiveness of the MOANA algorithm. The traditional ZDT benchmarks[30], as well as the 2019 CEC Multi-modal Multi-Objective benchmarks [33].Results of MOANA approach put to comparison with new multi-objective dragonfly algorithms MODA [22], multi-objective fitness-dependent optimizer MOFDO, MOPSO, and NSGA-III [24], and other contemporary approaches.

The ZDT Benchmark Functions (ZDT1 through ZDT6) are widely recognized for evaluating multi-objective evolutionary algorithms (MOEAs). These datasets present various challenges, including different Pareto front shapes convex, non-convex, and discontinuous. The ZDT functions primarily focus on two-objective optimization problems, enabling algorithms to be assessed on their ability to converge to the Pareto front while maintaining diversity in solutions. These benchmarks were chosen due to their simplicity, offering valuable insights into MOANA's core behaviors, particularly its ability to balance exploration and exploitation in standard optimization settings. In contrast, the CEC-2019 Multi-modal Multi-objective Benchmark represents a more complex and demanding set of multi-objective test functions. These benchmarks simulate real-world challenges, with multiple local Pareto fronts and varying decision spaces. Each function in the CEC-2019 series differs in terms of the number of objectives, constraints, and the complexity of the decision space. This makes the dataset particularly suited for evaluating MOANA’s capability to handle intricate, multi-modal landscapes. Some functions feature both global and local Pareto sets and their complexity scales with the number of decision variables and objectives. This makes the CEC-2019 dataset an ideal choice for testing MOANA's ability to navigate highly complex and multi-modal optimization tasks, further showcasing its robustness and applicability to real-world problems.

### *4.1 Results of the Classical ZDT Benchmarks*

The comparison of key parameter settings across the five algorithms—MOANA, MOFDO, MOPSO, NSGA-III, and MODA provides insights into the distinct configurations used for each. Table 1 highlights important aspects such as iterations, population size, repository size, crossover rates, grids per dimension, mutation rates, inflation rates, and the range of search parameters. This benchmark analysis, based on the ZDT1 through ZDT6 functions (refer to "Appendix B") See Table 9, which displays the mathematical definition, presents a challenging multi-objective optimization scenario, allowing for a thorough comparison of algorithm performance as seen in Table 2. The parameter settings for each algorithm are aligned with their original papers to ensure fairness in performance evaluation. After 100 iterations for each algorithm, with 500 agents in the population and repository, the results demonstrate the effectiveness of each approach, with MOANA's outcomes compared to other widely recognized algorithms[39] [22, 24, 43].The Figure 8 described The multi-objective algorithms achieved the best Pareto optimal front across several ZDT test functions. Can see Figure 8 in ("Appendix C")

**Table 2.** Parameter Settings for MOANA, MOFDO, MOPSO, NSGA-III, and MODA.

| Parameter | Algorithms MOANA | MOFDO | MOPSO | NSGA-III | MODA |
|---|---|---|---|---|---|
| Iterations | 100 | 100 | 100 | 100 | 200 |
| Population Size | 500 | 500 | 500 | 500 | 500 |
| Repository Size | 500 | 500 | 500 | 500 | 500 |
| Crossover | NaN | NaN | NaN | 0.5 | NaN |
| Number of Grids per Dimension | 7 | 14 | 7 | 10 | NaN |
| Polynomial Mutation Rate | 2.0 | 1/0.5 | 2.0 | 0.02 | 1/0.5 |
| Inflation rate | 0.1 | 0.1 | 0.2 | 0.1 | 0.1 |
| Rang | [-1,1] | [-1,1] | [0,1] | [-1,1] | [-1,1] |

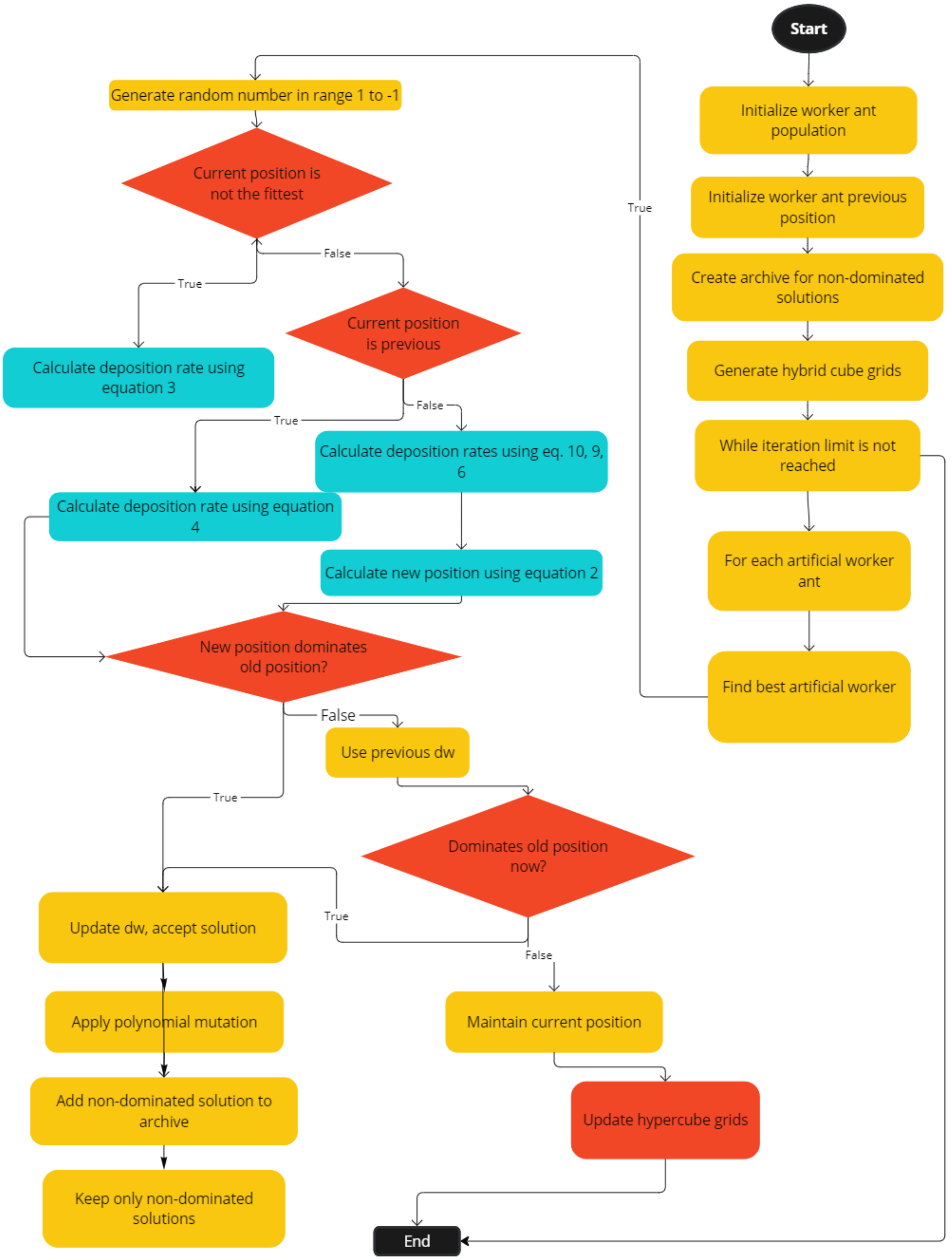
Start
Initialize worker ant population
Initialize worker ant previous position
Create archive for non-dominated solutions
Generate hybrid cube grids
While iteration limit is not reached
For each artificial worker ant
Find best artificial worker
True
Generate random number in range 1 to -1
Current position is not the fittest
False
True
Calculate deposition rate using equation 3
Current position is previous
False
True
Calculate deposition rates using eq. 10, 9, 6
Calculate deposition rate using equation 4
Calculate new position using equation 2
New position dominates old position?
False
Use previous dw
True
Dominates old position now?
True
False
Update dw, accept solution
Apply polynomial mutation
Add non-dominated solution to archive
Keep only non-dominated solutions
Maintain current position
Update hypercube grids
End

**Figure 2.** A Flowchart shows how MOANA works programmatically. The specified parameter configurations for MOANA are listed below: -The polynomial mutation rate is: 0.5 -The number of grids per dimension is 7- Remove Element is 2. - Best Ants Choice Element is: 2.

As evidence of performance of MOANA, the Figure 3. illustrates the solution landscape of the ZDT3 test function, demonstrating MOANA's effectiveness. In part (a), it can be observed that, initially, at iteration 5, only 387 Pareto front solutions are randomly distributed. However, as the iterations progress, as seen in parts (b), (c), and (d), MOANA successfully increases the number of solutions and refines their distribution towards the true Pareto front. The algorithm's progression from exploration to exploitation is evident in the increased convergence and density of solutions along the Pareto front. By the final iterations, the solutions effectively align with the Pareto front, confirming MOANA's strong performance in navigating the search space and optimizing multi-objective problems.

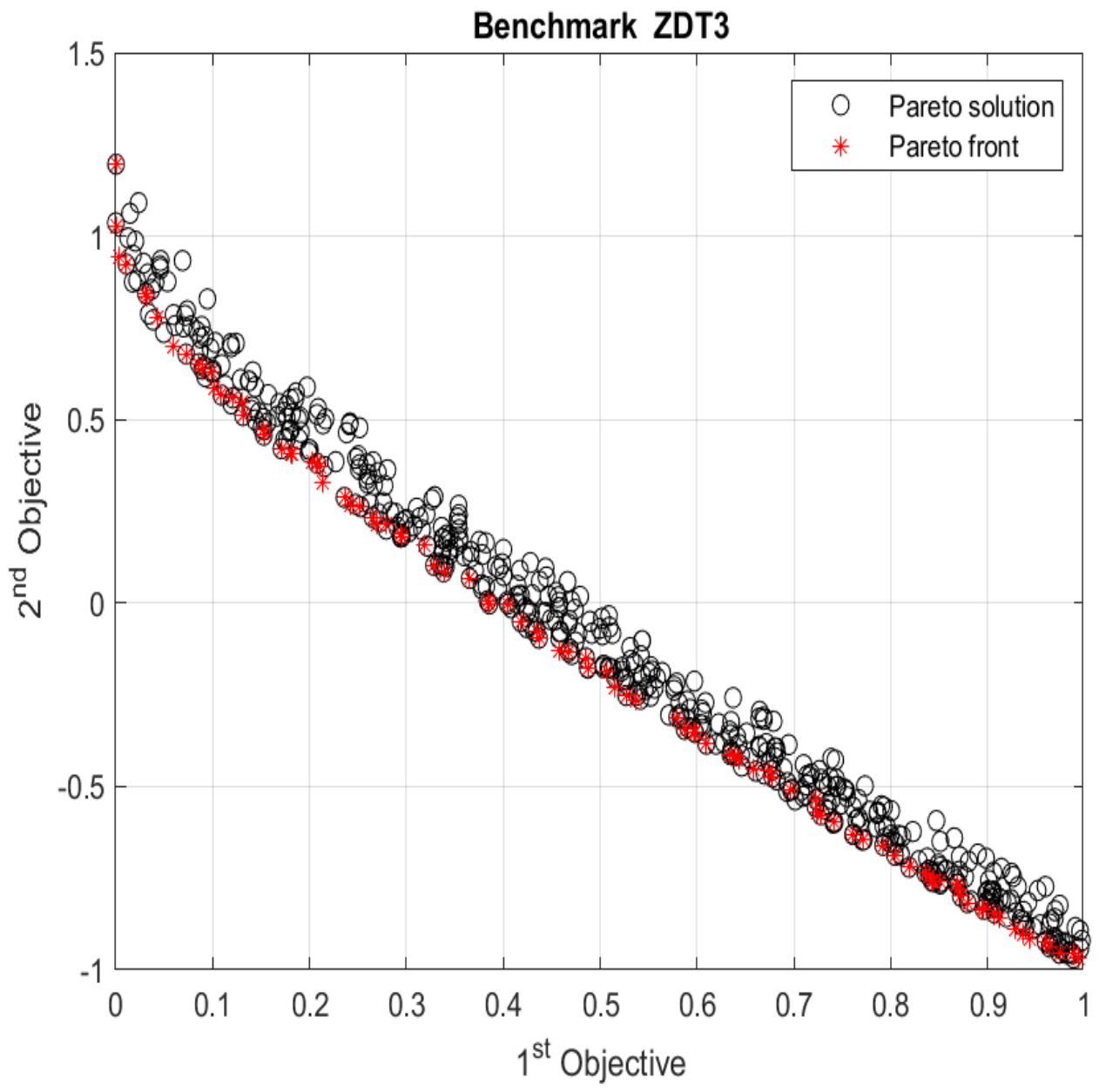

(a)

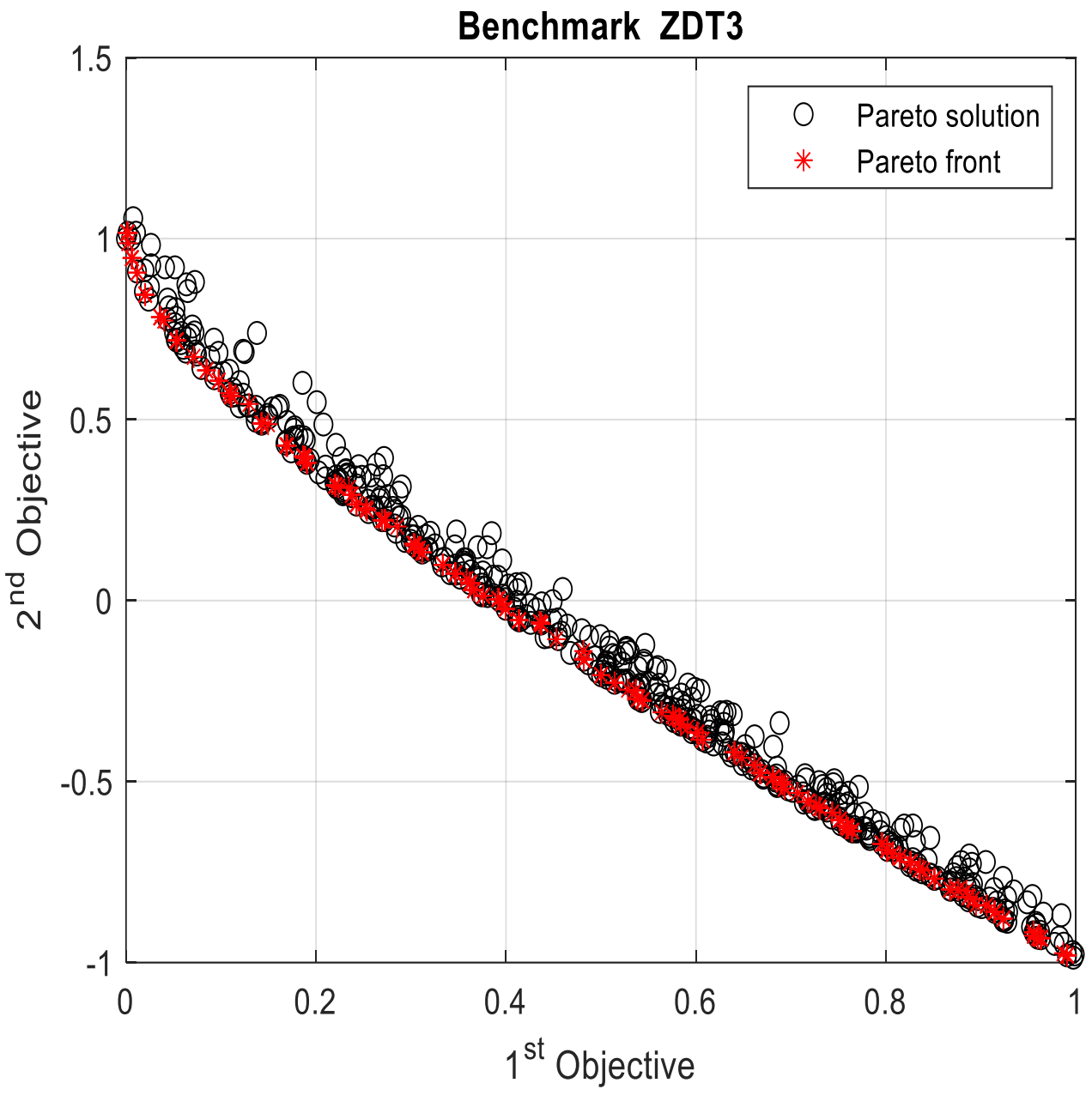

(b)

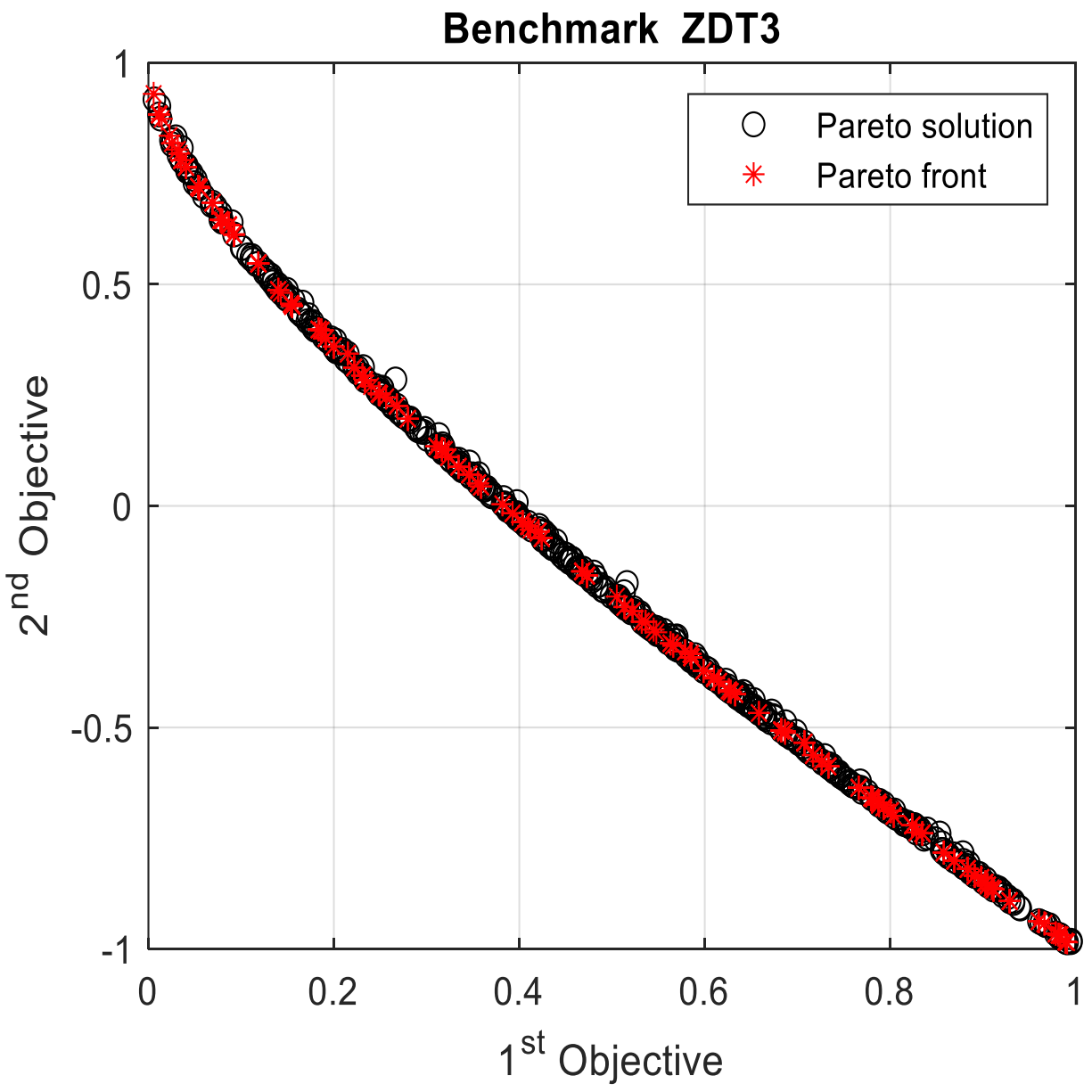

(c)

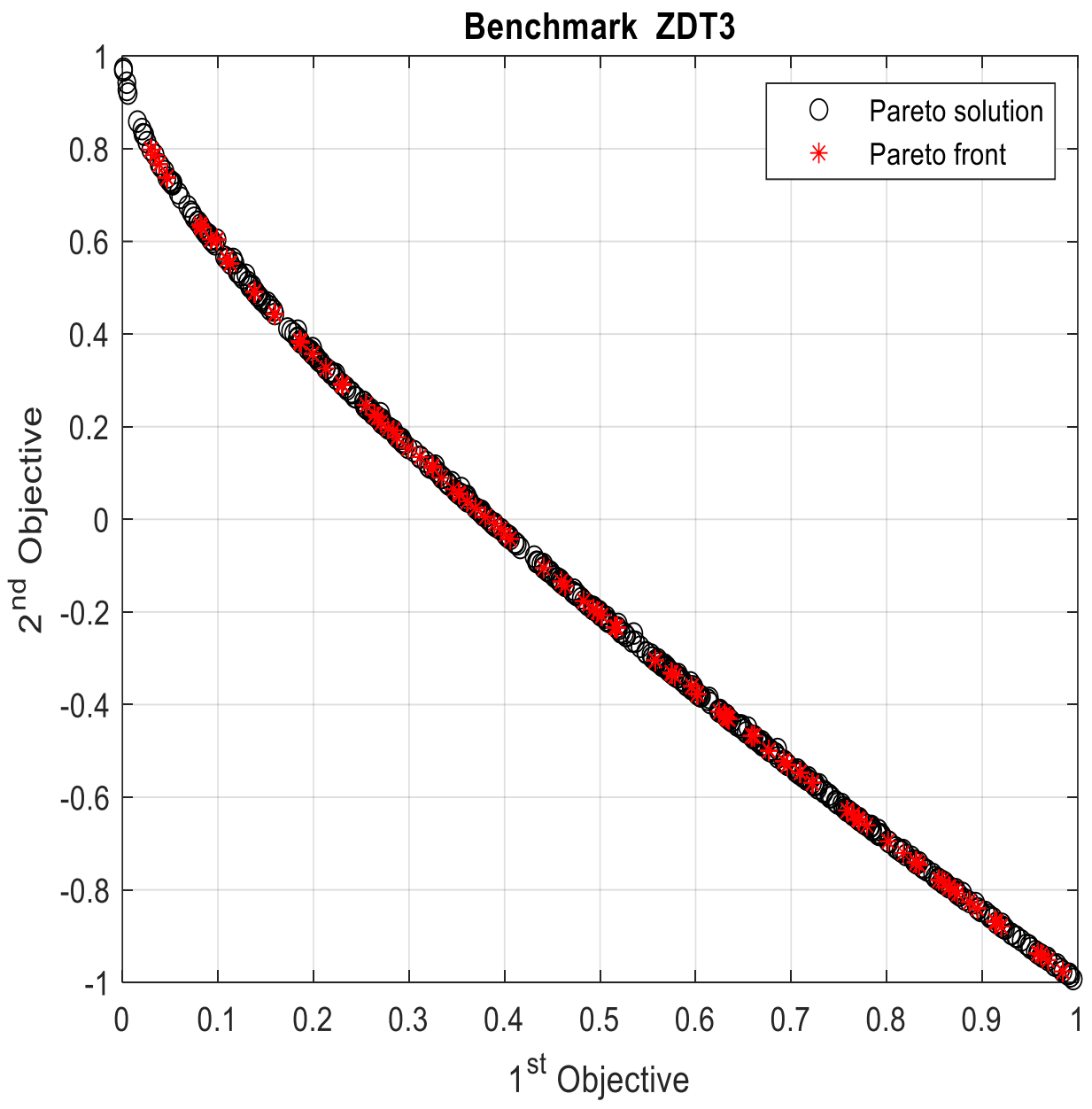

(d)

**Figure 3**. Demonstrate MOANA's process of solving ZDT3 test function by gradually improving an initially random solution until it reaches Pareto front optimality: (a)MOANA with 5 iterations, discovered 387 *PFs*;(b) MOANA with 20 iterations discovered 500 *PFs*;(d) MOANA with93 iterations, found 500 *PFs*;(c) MOANA with 200 iterations, found 500 *PFs*.

The inverse generational distance (IGD), which compares a solution set to a known Pareto front to evaluate its quality, is defined by the equation (12). In order to perform this comparison, the spatial separation between each component of the solution set and the Pareto front that the algorithm produces is quantified, as explained in the reference [44].

$$\mathrm{IGD} = \frac{\sqrt{\sum_{i=1}^{n} d_i^2}}{n} \tag{12}$$

The distance between *ith* real Pareto optimal solution in the collection of references and the nearest Pareto optimal solutions that have been discovered is known as the Euclidean distance, or *di*. The variable n denotes the aggregate count of genuine Pareto optimum options. When the value of IGD (Indicator for Generational Distance) is 0, It shows that every element that was produced is positioned exactly on the problem's actual Pareto front. The IGD was determined by collecting the results from individual runs for each method. Table 3 shows the calculated standard deviation, mean, worst, and best values of IGD.

To show the corresponding rankings of each algorithm in this context, ranking tables are employed. Table 4 demonstrates that MOANA achieved the top position in ZDT1, with a rank of 1. In ZDT2, MOANA also achieved a rank of 3, securing the third position. The cumulative total of all the ranks that a certain algorithm produces is known as the total rank. Making use of a ranking table is a straightforward way to demonstrate an algorithm's superiority over a set of rival algorithms. In addition, the Friedman test was utilized to ascertain the statistical significance of the data (refer to section 4.3.2). The ranking table was used for each test function. In Table4 illustrates, MOANA generally performs better than MOFDO, MOPSO MODA, and NSGA-III; yet it offers comparative outcomes and receives an overall ranking of 8.

**Table 3.** Results of the Traditional ZDT Benchmark

| Functions | Algorithms | IGD AVG | IGD STD | IGD best | IGD worst |
|---|---|---|---|---|---|
| **ZDT1** | MOANA | 0.0507 | 0.036041 | 0.023871 | 0.21846 |
| | MOFDO | 0.06758 | 0.030911 | 0.0018 | 2.61533 |
| | MOPSO | 0.56374 | 0.12618 | 0.32002 | 0.87601 |
| | NSGA-III | 15.0549 | 12.2435 | 0 | 32.3331 |
| | | 0.07653 | 0.012071 | 0.0420 | 0.59398 |
| **ZDT2** | MOANA | 0.016884 | 0.003756 | 0.009299 | 0.028032 |
| | MOFDO | 0.03511 | 0.00404 | 0.0207 | 0.0515 |
| | MOPSO | 0.33476 | 0.062539 | 0.20151 | 0.41251 |
| | NSGA-III | 0.54915 | 0.0548 | 0.0208 | 0.19889 |
| | MODA | 0.00292 | 0.00026 | 0.0002 | 0.0116 |
| **ZDT3** | MOANA | 0.051742 | 0.007879 | 0.030235 | 0.084617 |
| | MOFDO | 0.06676 | 0.023913 | 0.0014 | 2.2206 |
| | MOPSO | 0.93728 | 0.10293 | 0.60643 | 1.3493 |
| | NSGA-III | 16.2445 | 1.3133 | 0.83447 | 6.393 |
| | MODA | 0.07653 | 0.014411 | 0.0401 | 0.8267 |
| **ZDT4** | MOANA | 0.00145 | 0.011514 | 0 | 0.11454 |
| | MOFDO | 0.6802 | 0.352945 | 0.2679 | 1.6776 |
| | MOPSO | 1.2544 | 0.41974 | 0.28887 | 4.0945 |
| | NSGA-III | 173.4628 | 2.2739 | 32.288 | 39.9264 |
| | MODA | 64.9628 | 2.847807 | 51.742 | 500.93 |
| | MOANA | 0.39318 | 0.16227 | 0.047704 | 0.90018 |
| **ZDT6** | MOFDO | 0.35853 | 0.161795 | 0.1221 | 1.9125 |
| | MOPSO | 2.353 | 1.6138 | 0.49179 | 6.1428 |
| | NSGA-III | 6.08E+20 | 6.95E+18 | 0 | 6.08E+20 |
| | MODA | 0.11349 | 0.018270 | 0.0142 | 2.3938 |

**Table4.** The rankings table displays Table 3's algorithmic performances.

| Functions | MOANA ranking | MOFDO ranking | MPOSO ranking | NSGA-III ranking | MODA ranking |
|---|---|---|---|---|---|
| **ZDT1** | 1 | 2 | 4 | 5 | 3 |
| **ZDT2** | 2 | 3 | 5 | 4 | 1 |
| **ZDT3** | 1 | 2 | 4 | 5 | 3 |
| **ZDT4** | 1 | 2 | 3 | 4 | 5 |
| **ZDT6** | 3 | 2 | 5 | 5 | 1 |
| **Total sum** | 8 | 11 | 20 | 23 | 13 |

### *4.2 Results of CEC 2019 Multimodal multi-objective benchmarks*

According to [45], the mathematical description of the twelve CEC-2019 Multi-modal Multi-objective (MMO) benchmarks is provided in Table 10 (refer to "Appendix A"). This benchmark was selected because its test functions are more challenging compared to the ZDT benchmark used in MOANA. These functions present various characteristics, including issues related to varying PS and PF forms, both local and global PSs, as well as scalability in terms of the number of PSs, decision variables, and objectives the Figure 9. Described the performance of MOANA across several MMF CEC2019 test functions is illustrated. And the Figure 10. Described how the multi-objective algorithms achieved the best Pareto optimal front across several MMF CEC2019 test functions. Can see Figure 9 and Figure 10 in ("Appendix C").

As evidence of performance of MOANA, the Figure 4. illustrates the MMF4 benchmark solution landscape. In part (a), the initial random distribution of 360 Pareto front (*PF*) solutions is shown. As the iterations progress, as depicted in parts (b), (c), and (d), the number of (*PF)* solutions increases and becomes more evenly dispersed across the Pareto front. This demonstrates MOANA's ability to effectively explore the solution space and converge toward an optimal set of solutions. The increasing density and even distribution of solutions across the Pareto front reflect MOANA’s adaptive capability in maintaining diversity while efficiently guiding the search process toward global optima. Also showcases MOANA's effectiveness in exploring and exploiting the search space, particularly in managing complex, multi-modal optimization landscapes like MMF4, where multiple local Pareto fronts exist. MOANA's adaptive strategies allow it to discover a diverse set of solutions and successfully converge to the global Pareto front with each iteration. The Figure 4 also highlight how MOANA balances exploration (diversifying solutions) and exploitation (focusing on optimal regions) throughout the optimization process, demonstrating its ability to efficiently handle multi-objective optimization challenges. This balance contributes to MOANA’s superior performance in navigating complex search spaces and finding optimal solutions across multiple objectives.

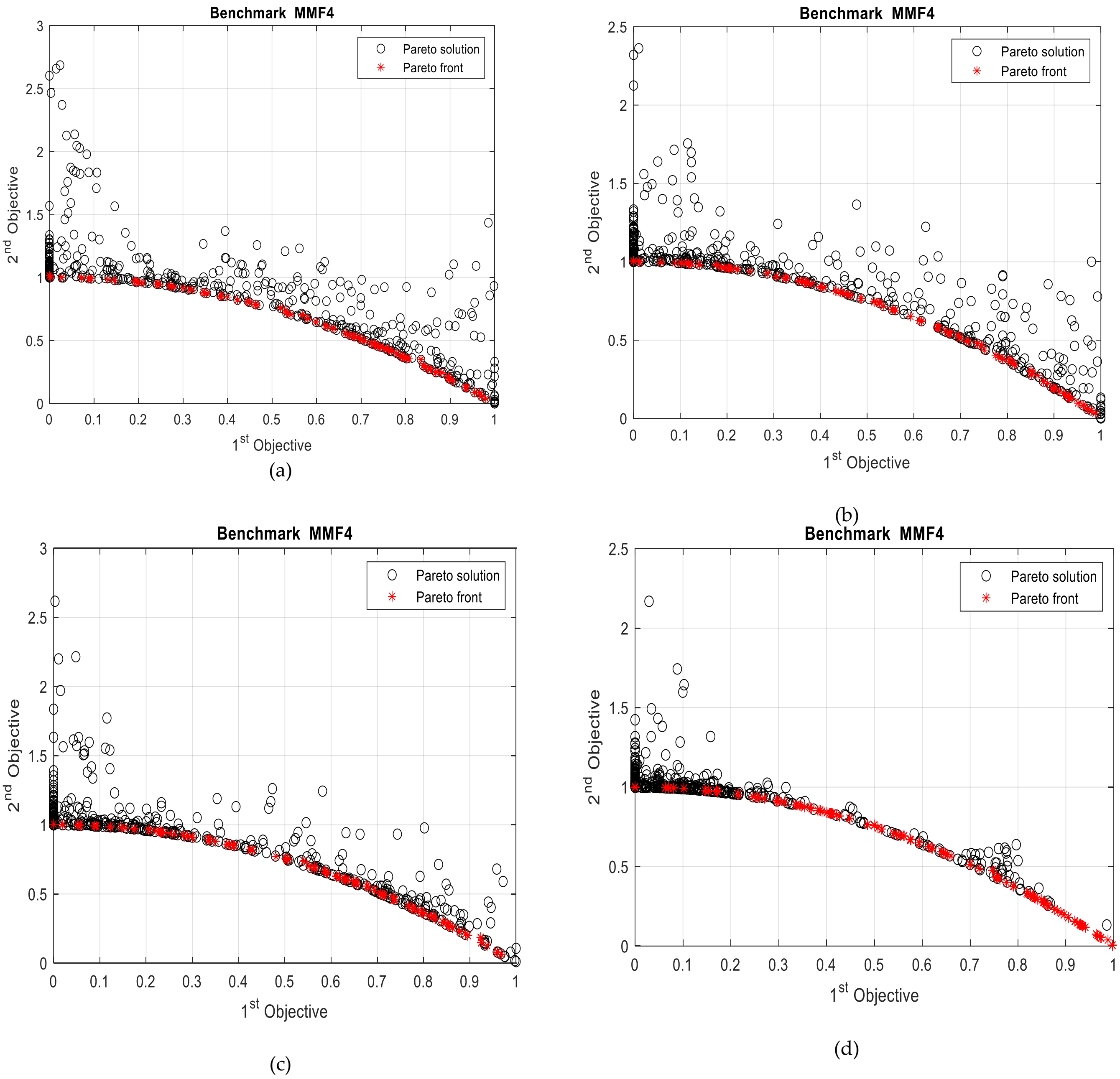

**Figure 4.** Demonstrate MOANA's process of solving MMF-4 test function by iteratively improving an initially random solution until reaching Pareto front optimality:(a) MOANA with 4 iterations, discovered 312 *PFs* ;(b) MOANA with 10 iterations, discovered 500 *PFs* ;(c) MOANA with 30 iterations, discovered 500 *PFs*;(d) MOANA with 167 iterations, discovered 500 *PFs*.

Table 5 provides a comparison of MOANA's results to those of MOFDO, MOPSO, NSGA-III, and MODA.
MOANA, ranked first and typically producing better outcomes, is summarized in Table 6, which presents the ranking system used to explain the results. MOFDO, MODA, MOPSO, and NSGA-III follow in second and third place, respectively.

**Table5.** The Results of Multi-Modal, Multi-Objective CEC 2019 Benchmarks

| Function | Algorithm | IGD AVG | IGD STD | IGD best |
|---|---|---|---|---|
| **MMF1** | MOANA | 0.10456 | 0.030503 | 0.076809 |
| | MOFDO | 0.18401921 | 0.0454458 | 0.0882267 |
| | MOPSO | 0.42858 | 0.14264 | 0.28265 |
| | NSGA-III | 7.0694 | 1.1494 | 0.018821 |
| | MODA | 0.87703300 | 0.5302916 | 0.3618665 |
| **MMF2** | MOANA | 0.046517 | 0.081073 | 0.021128 |
| | MOFDO | 0.09108902 | 0.0237087 | 0.0377645 |
| | MOPSO | 0.47075 | 0.17432 | 0.24579 |
| | NSGA-III | 5.95E+52 | 3.33E+53 | 0 |
| | MODA | 0.41152959 | 0.3041183 | 0.0883137 |
| **MMF3** | MOANA | 0.051916 | 0.028335 | 0.035151 |
| | MOFDO | 0.09121177 | 0.0184429 | 0.0412612 |
| | MOPSO | 0.55775 | 0.098921 | 0.34067 |
| | NSGA-III | 1.71E+53 | 1.71E+54 | 0 |
| | MODA | 0.38999723 | 0.3195349 | 0.0720422 |
| **MMF4** | MOANA | 0.025601 | 0.0067606 | 0.019853 |
| | MOFDO | 0.08195533 | 0.0340485 | 0.0453016 |
| | MOPSO | 0.54681 | 0.12362 | 0.42527 |
| | NSGA-III | 11.2789 | 2.2351 | 0.014653 |
| | MODA | 0.00781723 | 0.0038766 | 0.0003086 |
| **MMF5** | MOANA | 0.046148 | 0.016132 | 0.03013 |
| | MOFDO | 0.08166825 | 0.0203041 | 0.0410373 |
| | MOPSO | 0.4227 | 0.10278 | 0.25442 |
| | NSGA-III | 7.4504 | 1.2148 | 0.028032 |
| | MODA | 0.20697935 | 0.1068910 | 0.1177007 |
| **MMF6** | MOANA | 0.025749 | 0.016851 | 0.015363 |
| | MOFDO | 0.06319825 | 0.0052717 | 0.0435369 |
| | MOPSO | 0.36697 | 0.10461 | 0.26926 |
| | NSGA-III | 6.3483 | 1.1569 | 0.002903 |
| | MODA | 6.20722767 | 7.0529532 | 3.8844812 |
| **MMF7** | MOANA | 0.10082 | 0.017286 | 0.052247 |
| | MOFDO | 0.14853951 | 0.0219769 | 0.0870897 |
| | MOPSO | 0.26553 | 0.11412 | 0.13047 |
| | NSGA-III | 5.1419 | 0.87081 | 0.008045 |
| | MODA | 0.36139133 | 0.1036987 | 0.1322042 |
| **MMF8** | MOANA | 0.046957 | 0.074993 | 0.022752 |
| | MOFDO | 0.15550447 | 0.0869989 | 0.034329 |
| | MOPSO | 0.24122 | 0.056853 | 0.12072 |
| | NSGA-III | 0.01038634 | 0.0032172 | 0.0041821 |
| | MODA | 0.08058735 | 0.2652865 | 0.0056401 |
| **MMF9** | MOANA | 0.19218 | 0.046339 | 0.13157 |
| | MOFDO | 0.47321267 | 0.1219659 | 0.3404164 |
| | MOPSO | 1.33275589 | 0.142753 | 0.7792982 |
| | NSGA-III | 1.06E+00 | 8.89E-02 | 7.68E-05 |
| | MODA | 0.05060995 | 0.0274896 | 0.0051946 |
| **MMF10** | MOANA | 0.36877 | 0.15684 | 0.16607 |
| | MOFDO | 0.44207841 | 0.1277887 | 0.3104489 |
| | MOPSO | 1.00054897 | 0.1542964 | 0.7005662 |
| | NSGA-III | 3.89641261 | 4.6634273 | 0.002874 |
| | MODA | 0.09017605 | 0.0385574 | 0.0039308 |
| **MMF11** | MOANA | 0.39502 | 0.16976 | 0.2534 |
| | MOFDO | 0.09260275 | 0.0209854 | 0.0635536 |
| | MOPSO | 1.30789085 | 0.1622864 | 0.6847497 |
| | NSGA-III | 1.18058557 | 0.7034533 | 0.0034136 |
| | MODA | 0.09291338 | 0.0515551 | 0.0042148 |
| **MMF12** | MOANA | 0.034299 | 0.024017 | 0.019528 |
| | MOFDO | 0.08314653 | 0.0217281 | 0.0554114 |
| | MOPSO | 0.13651933 | 0.0237385 | 0.066709 |
| | NSGA-III | 0.35064339 | 0.1613096 | 0.3494061 |
| | MODA | 0.03661122 | 0.0119014 | 0.0035018 |

**Table6.** The rankings table displays Table 5's algorithmic performances

| Functions | MOANA ranking | MOFDO ranking | MPOSO ranking | NSGA-III ranking | MODA ranking |
|---|---|---|---|---|---|
| **MMF1** | 1 | 2 | 3 | 5 | 4 |
| **MMF2** | 1 | 2 | 4 | 5 | 3 |
| **MMF3** | 1 | 2 | 4 | 5 | 3 |
| **MMF4** | 2 | 3 | 4 | 5 | 1 |
| **MMF5** | 1 | 3 | 4 | 5 | 3 |
| **MMF6** | 1 | 2 | 3 | 4 | 5 |
| **MMF7** | 1 | 2 | 4 | 5 | 4 |
| **MMF8** | 3 | 4 | 2 | 1 | 5 |
| **MMF9** | 2 | 3 | 4 | 5 | 1 |
| **MMF10** | 2 | 3 | 4 | 5 | 1 |
| **MMF11** | 3 | 1 | 5 | 4 | 2 |
| **MMF12** | 1 | 3 | 4 | 5 | 2 |
| **Total sum** | 19 | 30 | 41 | 54 | 34 |

In this section, expanded the discussion on why MOANA performs well on some benchmarks, such as MMF4, while facing challenges on others, like MMF6, supported by statistical evidence from the Wilcoxon rank-sum test.

MMF4: Superior Performance of MOANA
MOANA consistently outperforms other algorithms, including MOFDO, MOPSO, and NSGA-III, on MMF4. This strong performance is primarily due to MOANA's adaptive balance between exploration and exploitation. The use of the deposition weight parameter allows the algorithm to effectively explore both dense and sparse regions of the search space, maintaining diversity while converging toward the Pareto front. Furthermore, the polynomial mutation mechanism plays a key role in helping MOANA escape local optima, which is particularly beneficial in multi-modal functions like MMF4, where multiple Pareto fronts exist. The statistical analysis confirms this advantage, with extremely low p-values (e.g., p = 6.654e-8), highlighting the algorithm’s significant superiority.

MMF6: Identifying Areas for Improvement
While MOANA generally shows strong results, its performance on MMF6 indicates room for improvement. MMF6 presents a complex landscape with numerous local optima, and while MOANA competes effectively with other algorithms, it doesn't dominate as clearly as it does in MMF4. The less pronounced difference in performance, especially compared to MODA, suggests that MOANA’s exploration strategy might not fully exploit the most promising regions in such rugged landscapes. Adjustments to the deposition weight and mutation strategy could enhance its ability to better navigate these challenging landscapes. The statistical analysis supports this observation, as the p-value (p = 0.09609 against MOFDO) is statistically significant but not as low as in other cases, indicating that MOANA’s advantage is less substantial here.

The Wilcoxon rank-sum test (in section 4.3.1) reinforces these findings, confirming that MOANA performs significantly better on benchmarks like MMF4, where its adaptive mechanisms are most effective. However, in functions like MMF6, while MOANA still outperforms many algorithms, the statistical differences are less pronounced, pointing to areas for further algorithmic refinement.

### *4.3 The Statistical Analysis*

In this section, we present a detailed statistical analysis to evaluate the performance of the Multi-Objective Ant Nesting Algorithm (MOANA) in comparison to other well-established optimization algorithms, including MOFDO, MOPSO, NSGA-III, and MODA. Statistical tests are essential for ensuring the robustness of the comparative results, moving beyond basic performance metrics to establish the significance of the differences observed between these algorithms.

The analysis applies non-parametric statistical tests, such as the Mann-Whitney U test, Wilcoxon rank-sum test, and Friedman test, which are appropriate for comparing performance across various problem instances, especially when the data distribution is not normal. We compute p-values, test statistics, and confidence intervals for each comparison to assess whether the observed performance differences between MOANA and the competing algorithms are statistically significant. This rigorous approach validates the results, ensuring that MOANA's performance advantages are not due to random variations but reflect genuine improvements in algorithmic efficiency and solution quality.

The following subsections provide detailed insights into the statistical results, explaining their significance and how they relate to MOANA's effectiveness in solving multi-objective optimization problems.

#### *4.3.1 The Mann-Whitney U test and the Wilcoxon rank-sum*

The Wilcoxon rank-sum test demonstrates that MOANA consistently performs well across most ZDT and MMF functions, significantly outperforming traditional algorithms like MOFDO, MOPSO, and NSGA-III in terms of convergence and diversity. However, in some cases, such as ZDT6 and MMF7, the differences are less pronounced, as shown in Table 7 and Table 8.

To verify whether the results in Tables 3 and 5 are statistically significant, the Wilcoxon rank-sum test was conducted to calculate the p-values comparing MOANA with other algorithms. As presented in Table 7 and Table 8, the majority of the results in Table 3 (ZDT benchmark results) and Table 5 (CEC 2019 benchmark results) are statistically significant, as the p-values are smaller than 0.05.

**Table 7.** The Wilcoxon Rank-Sum Test (P-Value) For ZDT Benchmarks

| Functions | MOANA Vs. MOFDO | MOANA Vs. MPOSO | MOANA Vs. NSGA-III | MOANA Vs. MODA |
|---|---|---|---|---|
| **ZDT1** | 0.001641 | 0.0001172 | 0.00007057 | 0.000003239 |
| **ZDT2** | 0.09322 | 8.061e-10 | 1.075e-9 | 1.863e-9 |
| **ZDT3** | 1.863e-9 | 0.0008054 | 0.04338 | 0.4771 |
| **ZDT4** | 0.0002201 | 0.04603 | 0.002782 | 8.451e-10 |
| **ZDT6** | 0.4161 | 0.000006918 | 0.166 | 0.0008044 |

**ZDT1:** MOANA consistently performs better than all other algorithms, with very low p-values, indicating a statistically significant improvement.

**ZDT2:** MOANA shows a strong advantage over MOPSO, NSGA-III, and MODA, but the difference with MOFDO is not statistically significant (p = 0.09322).

**ZDT3:** MOANA performs well against MOFDO and MOPSO, but the differences with NSGA-III and MODA are less pronounced, as indicated by higher p-values.

**ZDT4:** MOANA significantly outperforms all algorithms, including MOFDO, MOPSO, NSGA-III, and MODA.

**ZDT6:** While MOANA performs significantly better than MOPSO and MODA, the differences with MOFDO and NSGA-III are not statistically significant.

**Table 8.** The Wilcoxon Rank-Sum Test (P-Value) For Multi-Modal, Multi-Objective CEC 2019 Benchmarks

| Functions | MOANA Vs. MOFDO | MOANA Vs. MPOSO | MOANA Vs. NSGA-III | MOANA Vs. MODA |
|---|---|---|---|---|
| **MMF1** | 0.006195 | 0.01853 | 1.863e-9 | 0.00004408 |
| **MMF2** | 0.0002833 | 0.03655 | 1.365e-10 | 0.0002091 |
| **MMF3** | 0.3494 | 0.144 | 8.986e-8 | 0.00003031 |
| **MMF4** | 0.02963 | 0.01846 | 9.313e-10 | 6.654e-8 |
| **MMF5** | 0.000002722 | 0.02304 | 0.01641 | 0.005489 |
| **MMF6** | 0.09609 | 0.00001273 | 0.00154 | 0.0000158 |
| **MMF7** | 0.3858 | 0.2227 | 0.02114 | 0.07135 |
| **MMF8** | 0.0001987 | 0.0001507 | 0.03059 | 3.749e-7 |
| **MMF9** | 0.09092 | 0.004944 | 0.09069 | 0.1125 |
| **MMF10** | 0.1214 | 0.6552 | 0.001532 | 0.001121 |
| **MMF11** | 0.06753 | 0.0001608 | 0.6168 | 0.4277 |
| **MMF12** | 0.3265 | 0.418 | 0.01367 | 0.248 |

MOANA performs significantly better than NSGA-III and MODA in most functions. For functions like MMF1, MMF2, MMF4, MMF5, and MMF6, MOANA shows significant improvements over all algorithms.
However, for some functions like MMF3, MMF7, and MMF9, the differences between MOANA and MOFDO/MOPSO are not statistically significant.

### 4.3.2 *The results of the Friedman test*

The Friedman test demonstrated that the results were statistically significant. This test determines whether three or more related samples have substantial differences. It serves as the non-parametric equivalent of repeated measures analysis of variance. The Friedman test was used to assess the statistical significance of the findings presented in Table 3 and Table 5. There are two hypotheses in the Friedman test: the null hypothesis asserts that there are no significant differences between dependent groups, while the alternative hypothesis suggests that a significant difference exists between the groups. Rather than using actual values, the Friedman test relies on the ranks of the data. The equation is used to compute the Friedman test as see in equation (13).

$$x_r^2 = \frac{12}{nk(k+1)} \sum R^2 - 3n(k+1) \tag{13}$$

To compute a Friedman test, one must establish the state decision rule by utilizing the supplied formula. (n) is the multiple test function s, (k) is the number of groups (i.e., the number of the compared algorithms), (R) represents the square root of the overall rank of every group, and (x2r) is the Chi-square statistic. Computed by summing the rank values from Tables 4 and 6. To determine the degree of freedom, it is necessary to deduct 1 from the total number of groups, which in this particular instance is 5. This calculation yields a value of df = k - 1 = 5 - 1 = 4. To get the decision rule state for a given significance level alpha, we can consult the Chi-squared distribution table. In this case, the degree of freedom (df) is 4, and the corresponding value is 9.488 for a p-value less than 0.05 [30]. The computation of the Friedman test for the combined Table 4 and Table 6 will be conducted. The overall rankings for the algorithms R are as follows: MOANA = 27 MOFDO = 41, MODA = 47, MOPSO = 61, and NSGA-III = 77. The value of n is 17, which corresponds to the number of test function results present in both Tables 2 and 4. Additionally, the value of k is:

$$x_r^2 = \frac{12}{17 * 5(5+1)} \sum(27^2 + 41^2 + 61^2 + 47^2 + 77^2) - 3 * 17 * (5+1)$$

The ($x_r^2 = 24.8.627$) and p-value is($7.45 * 10^{-205}$). Null hypothesis is rejected by Friedman test, indicating that the data are statistically significant with a p-value of (<0.5). A significantly low p-value indicates the presence of substantial differences between the groups. It provides strong evidence against the null hypothesis and establishes a precise statistical basis for the observed effects or differences by proving that observed differences are not a result of the random variation. This strengthens the case for the existence of significant and genuine phenomena in the community under investigation.

### *4.4 Engineering design application*

The application of MOANA in optimizing the design problem of a welded beam is showcased, followed by a thorough analysis of the problem specifications and the findings obtained. The welded beam design problem is a well-known engineering challenge that has been extensively studied by researchers as a benchmark for evaluating various multi-objective algorithms[46],[47]. Figure5. illustrates this problem, which involves four real-valued variables: *x = (b, t, l, h)*. In this context, *b* represents the horizontal dimension of the beam, *t* denotes the vertical dimension, *l* indicates the welds' length, and *h* refers to the welds' thickness. *P* represents the amount of load applied to the beam. This design problem involves a bi-objective optimization to be minimized, as shown in Equation (14). The two objectives are conflicting in nature: the first objective is to minimize the cost of fabrication (measured on the X-axis in currency), and the second objective is to reduce the end deflection of the welded beam (measured on the Y-axis in meters). The goal is to optimize these objectives simultaneously.

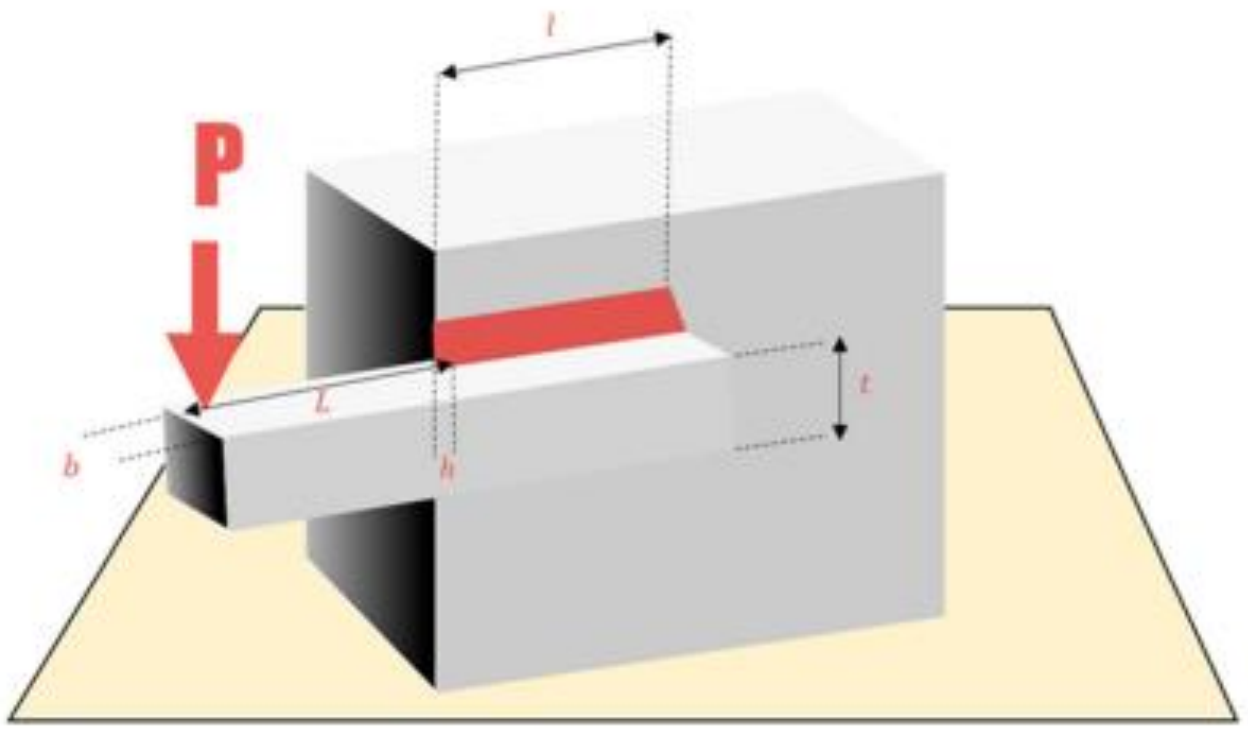

**Figure 5.** The problem of designing a welded beam

Equation (14) shows that there are four constraints in this situation that must be taken into account. A design that is not satisfactory will arise from exceeding these limitations. The primary restriction is to keep the shear stress generated at support site of the beam below the permitted threshold of (13,600psi). The second restriction is to keep the normal stress generated at the support point of beam below the material's allowable yield strength of 30,000 psi. The final restriction is to make sure that, taking into account practical concerns, the beam's width and the weld's width match.

$$
\begin{aligned}
&\text{Minimize } f1(\vec{x}) = 1.10471h^2 l + 0.04811tb(14.0 + l),\\
&\text{Minimize } f2(\vec{x}) = \frac{2.1952}{t^3 b},\\
&\text{Subject to:}\\
&g1(\vec{x}) \equiv 13{,}600 - T(\vec{x}) \geq 0,\\
&g2(\vec{x}) \equiv 13{,}600 - \sigma(\vec{x}) \geq 0,\\
&g3(\vec{x}) \equiv b - h \geq 0,\\
&g4(\vec{x}) \equiv Pc(\vec{x}) - 6000 \geq 0,\\
&0.125 \leq h, b \leq 5.0\\
&0.1 \leq l, t \leq 10.0 \qquad (14)
\end{aligned}
$$

Constraint 4 ensures that the beam's buckling load, $pc(\vec{x})$exceeds the applied force, F, which is precisely 6000 pounds. The shear stress $T(\vec{x})$ and the buckling load $\curlyvee(\vec{x})$ can be estimated by using Equations (15) & (16) respectively.

$$
\begin{aligned}
T(\vec{x}) &= \sqrt{(T')^2 + (T'')^2 + (lT'T'')/\sqrt{0.25(l^2 + (h+t)^2)}}\\
T' &= \frac{6000}{\sqrt{2}hl}\\
T'' &= \frac{6000(14 + 0.5l)\sqrt{0.25(l^2 + (h+t)^2)}}{2\left\{0.707\, hl\left(\frac{l^2}{12} + 0.25(h+t)^2\right)\right\}}\\
\curlyvee(\vec{x}) &= \frac{504000}{t^2 b} \qquad (15)\\
\text{Pc}\,(\vec{x}) &= 64746.022(1 - 0.0282346t)tb^3 \qquad (16)
\end{aligned}
$$

The design problem of the welded beam is optimized with the MOANA approach. The technique is utilized for solving such engineering design problems for a total of 100 iterations, employing 100 search agents. An achievement of size 100 is used to store the Pareto front solutions (Pfs). As shown in Figure (6). the obtained Pfs are split equally between the two goals of Deflection and Cost. Furthermore, the majority of these Pfs are situated directly on or close to the genuine Pfs that are documented in the literature[48]. Furthermore, MOANA offers a diverse range of viable options for decision-makers to select from. The extensive selection and efficient distribution of the achieved Pfs demonstrates the advanced proficiency of MOANA in effectively addressing real-world engineering design challenges.

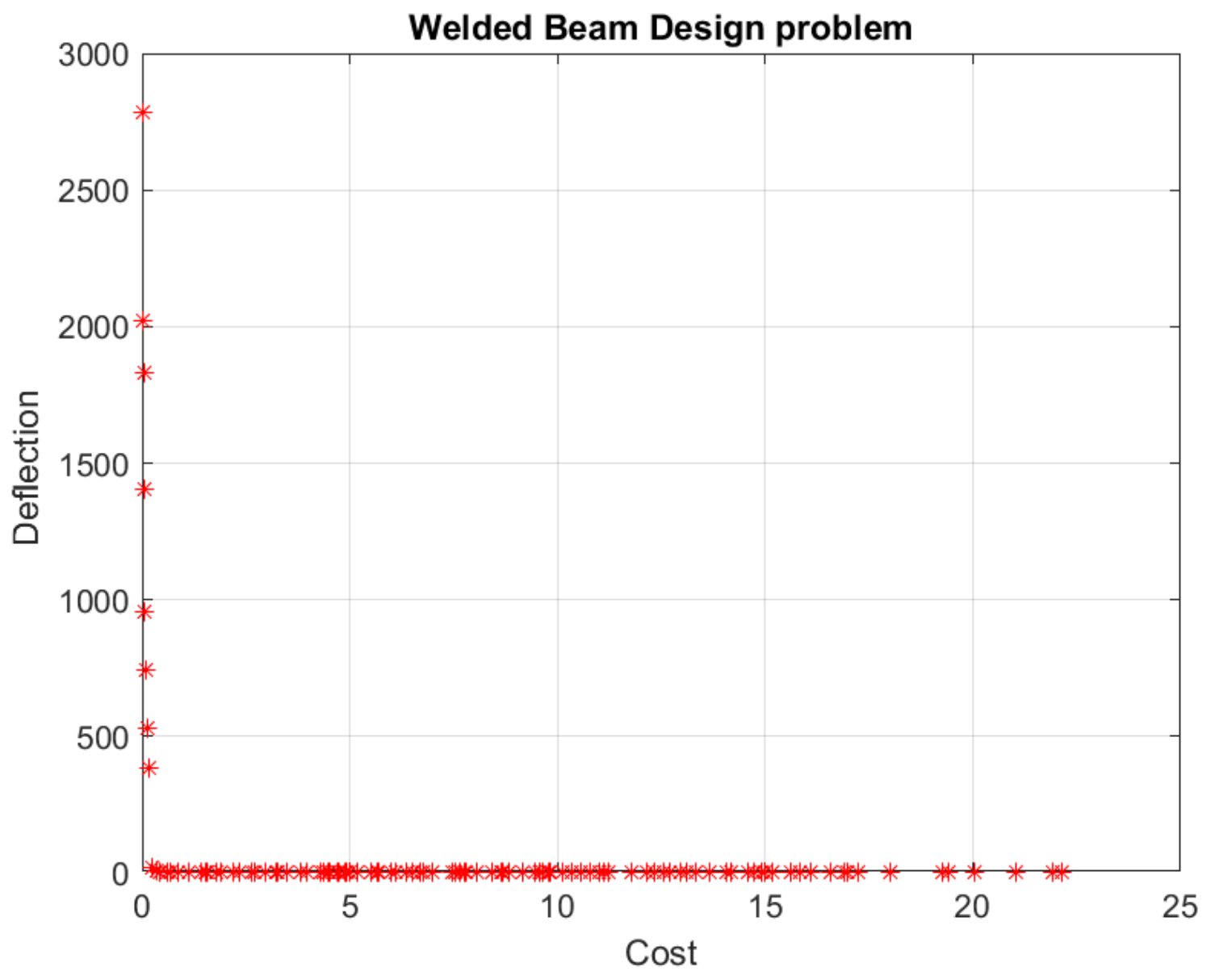

**Figure 6.** Results from MOANA for problems of designed welded beams

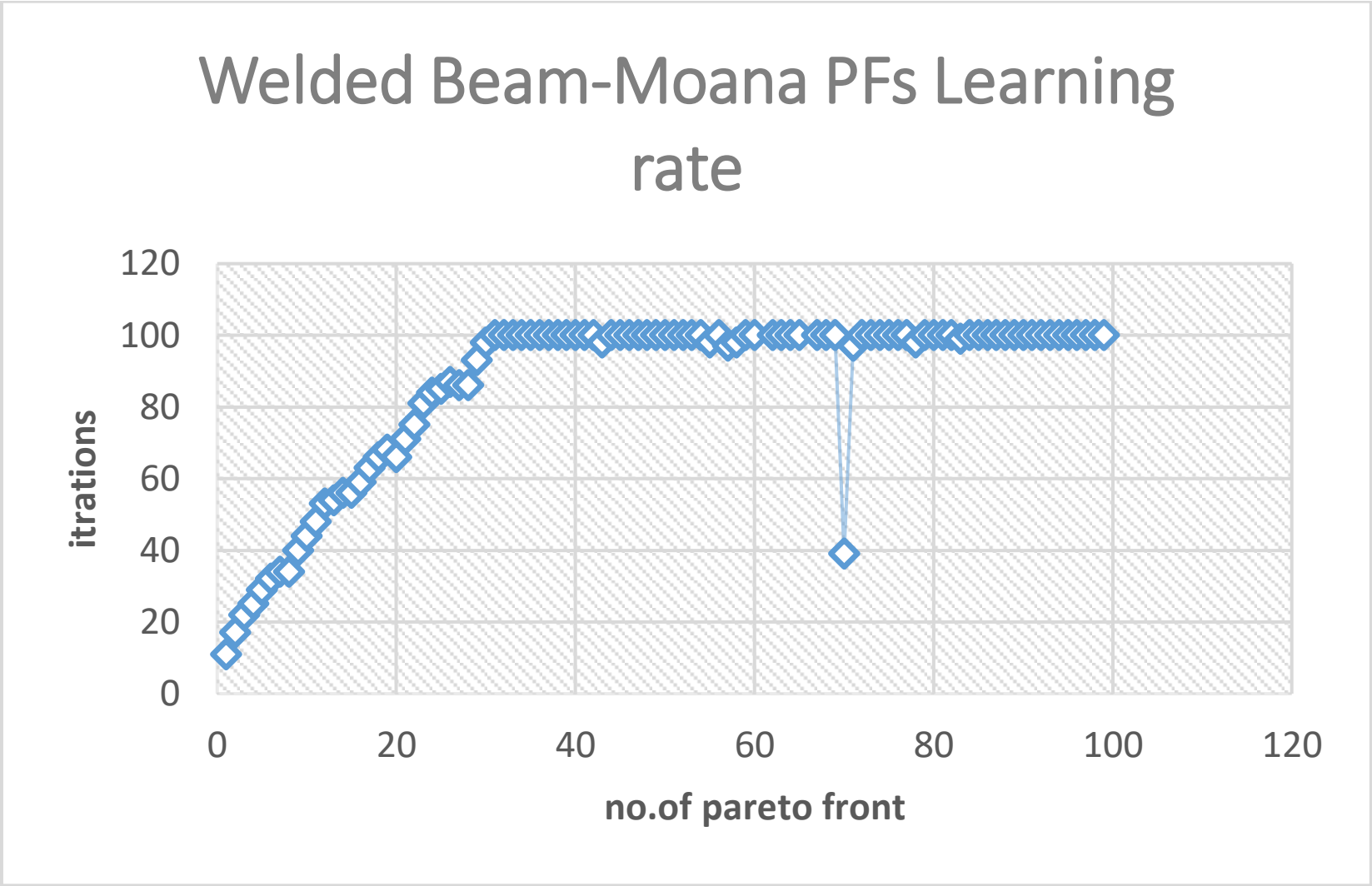

**Figure 7.** Shows MOANA Pfs discovering rate

The Pfs are saved in archives with a capacity of 100. Figure (7). demonstrates that the acquired Pfs are split equally between the two goals, namely Deflection and Cost are predominantly situated. Adjacent to or close to the verified Pfs documented in the literature [40].Regarding the rate at which MOANA discovers Pfs, it begins with 11 Pfs in the first iteration and then rapidly increases to 100 Pfs by the 31 iteration, as shown in Figure (7). This demonstrates how the algorithm effectively enhances several initial solutions towards achieving optimality, resulting in certain Pfs becoming non-dominated solutions intermittently. Consequently, they are eliminated from the archive. This phenomenon is evident in iterations 31-100 depicted in Figure 6, whereby the discovery rate of Pfs initiates with a modest quantity and progressively escalates until it attains the maximum capacity of the archive. MOANA consistently improves all solutions, whether they are dominated or non-dominated, in each iteration. By possessing this attribute, MOANA is able to steer clear of localized solutions and finally attain optimality[49].

## 5. Discussion on the Limitations of MOANA

Despite the promising performance of the Multi-Objective Ant Nesting Algorithm (MOANA), there are certain limitations, particularly concerning scalability, computational complexity, and convergence speed when tackling highly complex problems.

1. Scalability:
   One of the primary challenges for MOANA is its scalability when applied to large-scale multi-objective optimization problems. As the number of objectives or decision variables increases, the computational resources required to maintain an extensive Pareto front and effectively explore the solution space grow significantly. The hypercube grid-based approach used by MOANA, while beneficial for maintaining solution diversity, can become computationally expensive in higher dimensions, leading to slower performance as the problem size increases.

2. Computational Complexity:
   The algorithm's complexity is directly influenced by the need to calculate and compare solutions to maintain the Pareto front. MOANA's use of deposition weights, polynomial mutation, and grid partitioning strategy, while effective in maintaining diversity and improving exploration, contribute to the overall computational burden. The complexity of these operations can make the algorithm slower compared to other, more simplified multi-objective optimization algorithms. Moreover, MOANA's adaptive mechanisms, such as adjusting deposition rates based on a solution's position within the grid, require frequent recalculations and comparisons across solutions. This increases the computational load, particularly for problems with a large number of local Pareto fronts or highly rugged search spaces.

3. Convergence Speed
   MOANA's convergence speed can be slower on highly complex, multi-modal landscapes with numerous local optima. While MOANA's adaptive exploration-exploitation mechanism helps avoid local optima in many cases, it might not fully exploit the most promising regions efficiently, leading to slower convergence compared to algorithms designed specifically for faster convergence, such as differential evolution or covariance matrix adaptation strategies. For problems where the global Pareto front is difficult to identify, MOANA may require a large number of iterations to adequately explore and converge on the true Pareto-optimal solutions. This could result in longer runtimes, particularly in real-world applications with strict time or resource constraints.

## 6. Future Research Directions:

To address the current limitations, future research could explore hybridizing MOANA with other metaheuristic approaches to enhance its scalability and convergence speed. Techniques such as parallel computing, adaptive population sizing, or incorporating surrogate models could help reduce the computational overhead and improve performance on large-scale and highly complex problems. Additionally, further tuning of parameters like deposition weight and mutation rates for specific problem types could lead to more efficient convergence in complex multi-modal landscapes. While MOANA has been applied to standard benchmark functions and real-world problems like welded beam design, there is significant potential for extending its application to large-scale, multi-objective optimization problems in engineering. These could include the design of pressure vessels, speed reducers, automotive side-impact protection systems, coil compression springs, and four-bar truss problems. Future work should also focus on extending MOANA to tackle higher-dimensional problems and integrating it with other optimization techniques to further enhance its performance. Applying MOANA to more diverse and complex real-world problems would validate its versatility and effectiveness across various domains, such as logistics, healthcare, industrial design, supply chain management, transportation, and energy systems, and could offer valuable insights into its performance in practical settings. This could lead to further refinements and enhancements, ultimately making MOANA a more robust solution for real-world, multi-objective optimization challenges.

## 7. Conclusion

This study introduced MOANA, a novel multi-objective optimization algorithm inspired by the ant nesting process and adapted from the single-objective ANA. By integrating topographical, historical, and situational knowledge, MOANA provides a robust, population-based meta-heuristic approach to solving complex optimization problems. Its performance was evaluated using both traditional ZDT benchmarks and the more challenging CEC 2019 multi-modal benchmark, where it was compared against MOPSO, MODA, NSGA-III, and MOFDO. MOANA consistently demonstrated superior performance in most cases and matched the results of other algorithms in others. The performance differences between MOANA and other algorithms were confirmed through rigorous statistical analysis, including the Wilcoxon rank-sum test and the Friedman test. These tests provided robust evidence that the observed improvements in MOANA’s results were statistically significant, especially in functions like ZDT1 and MMF4, where MOANA significantly outperformed the competing algorithms. The p-values and rankings tables in the results section further highlight MOANA’s effectiveness in balancing exploration and exploitation, allowing it to efficiently converge to the global Pareto front while maintaining diversity in the solution set. When applied to the welded beam design problem, MOANA effectively delivered diverse and evenly distributed Pareto optimal solutions, offering decision-makers a broader range of options. MOANA's adaptive mechanisms, such as its use of the deposition weight (dw) parameter, polynomial mutation, and Hypercube grids for guide selection, were key to achieving rapid convergence while maintaining diversity, especially in complex landscapes like those in the CEC 2019 benchmark.
However, MOANA has limitations, particularly in terms of computational complexity and scalability for larger, high-dimensional problems. These challenges present opportunities for future research, such as optimizing MOANA’s parameters, enhancing learning mechanisms, and hybridizing it with other algorithms to improve its convergence speed and efficiency in large-scale applications.

**Author Contributions:** Conceptualization, N.A.R; methodology, N.A.R.; software, N.A.R.; validation, N.A.R; formal analysis N.A.R, Y.A.R, T.A.R.; investigation, Y.A.R, T.A.R; resources, N.A.R, Y.A.R, T.A.R.; data curation, N.A.R.; writing—original draft preparation, N.A.R.; writing—review and editing N.A.R, Y.A.R, T.A.R; visualization, N.A.R.; supervision, Y.A.R, T.A.R.; project administration, N.A.R, Y.A.R, T.A.R.; funding acquisition, Y.A.R. All authors have read and agreed to the published version of the manuscript.

**Funding:** This research received internal funding.

**Conflicts of Interest:** The **authors** declare no conflicts of interest.

**Data Availability Statement:** The datasets generated during and/or analyzed during the current study are available from the corresponding author upon reasonable request.

**Appendix A**

**Table 9.** ZDT benchmark Mathematical definition.

| Functions | Mathematical definition |
| --- | --- |
| ZDT1 | $g(x) = 1 + 9(\sum_{i=2}^{n} x_i)/(n-1)$<br>$F_1(x) = x_1$<br>$F_2(x) = g(x)\left[1 - \sqrt{x_1/g(x)}\right] x \in [0,1]$ |
| ZDT2 | $g(x) = 1 + 9(\sum_{i=2}^{n} x_i)/(n-1)$<br>$F_1(x) = x_1$<br>$F_2(x) = g(x)[1 - (x_1/g(x))^2] x \in [0,1]$ |
| ZDT3 | $g(x) = 1 + 9(\sum_{i=2}^{n} x_i)/(n-1)$<br>$F_1(x) = x_1$<br>$F_2(x) = g(x)\left[1 - \sqrt{x_1/g(x)} - x_1/g(x)\sin(10\pi x_1)\right] x \in [0,1]$. |
| ZDT4 | $g(x) = 91 + \sum_{i=2}^{n} [x_i^2 - 10\cos(4\pi x_i)]$<br>$F_1(x) = x_1$<br>$F_2(x) = g(x)\left[1 - \sqrt{x_1/g(x)}\right] x_1 \in [0,1], x_i \in [-5,5] i = 2, \cdots, 10$. |
| ZDT6 | $g(x) = 1 + 9[(\sum_{i=2}^{n} x_i)/(n-1)]^{0.25}$<br>$F_1(x) = 1 - \exp(-4x_1)\sin^6(6\pi x_1)$<br>$F_2(x) = g(x)[1 - (f_1(x)/g(x))^2] x \in [0,1]$ |

## Appendix B

**Table 10**. CEC 2019 multimodal multi-objective benchmark mathematical definition [37].

| Functions | Mathematical definitions | Range |
|---|---|---|
| MMF1 | $\begin{cases} f_1 = \lvert x_1 - 2 \rvert \\ f_2 = 1 - \sqrt{\lvert x_1 - 2 \rvert} + 2\left(x_2 - \sin\left(6\pi \lvert x_1 - 2 \rvert + \pi\right)\right)^2 \end{cases}$ | $x_1 \in [1,3]$, $x_2 \in [-1,1]$ |
| MMF2 | $\begin{cases} f_1 = x_1 \\ f_2 = \begin{cases} 1-\sqrt{x_1}+2\left(4\left(x_2-\sqrt{x_1}\right)-^2 \\ 2\cos\left(\frac{20\left(x_2-\sqrt{x_1}\right)\pi}{\sqrt{2}}\right)+2\right),\ 0 \le x_2 \le 1 \\ 1-\sqrt{x_1}+2\left(4\left(x_2-1-\sqrt{x_1}\right)-^2 \\ \cos\left(\frac{20\left(x_2-1-\sqrt{x_1}\right)^2\pi}{\sqrt{2}}\right)+2\right), 1 < x_2 \le 2 \end{cases} \end{cases}$ | $x_1 \in [0,1],\ x_2 \in [0,2]$ |
| MMF3 | $\begin{cases} f_1 = x_1 \\ f_2 = \begin{cases} 1-\sqrt{x_1}+2\left(4\left(x_2-\sqrt{x_1}\right)-^2 \\ 2\cos\left(\frac{20\left(x_2-\sqrt{x_1}\right)\pi}{\sqrt{2}}\right)+2\right) \\ ,\ 0 \le x_2 \le 0.5, 0.5 < x_2 < 1 \quad 0.25 < x_1 \le 1 \\ 1-\sqrt{x_1}+2\left(4\left(x_2-0.5-\sqrt{x_1}\right)^2 \\ -\cos\left(\frac{20\left(x_2-0.5-\sqrt{x_1}\right)^2\pi}{\sqrt{2}}\right)+2\right) \\ , 1 \le x_2 \le 1.5, 0 \le x_1 < 0.25 \quad 0.5 < x_2 < 1 \end{cases} \end{cases}$ | $x_1 \in [0,1],\ x_2 \in [0,1.5]$ |
| MMF4 | $\begin{cases} f_1 = x_1 \\ f_2 = \begin{cases} \begin{cases} 1-x_1^2+2\left(x_2-\sin\left(\pi\lvert x_1\rvert\right)\right)^2 & 0 \le x_2 < 1 \\ 1-x_1^2+2\left(x_2-1-\sin\left(\pi\lvert x_1\rvert\right)\right)^2 & 1 \le x_2 \le 2 \end{cases} \end{cases} \end{cases}$ | $x_1 \in [-1,1], x_2 \in [0,2]$ |
| MMF 5 | $\begin{cases} f_1 = \lvert x_1 - 2 \rvert \\ f_2 = \begin{cases} 1 - \sqrt{\lvert x_1 - 2 \rvert} + 2\left(x_2 - \sin\left(6\pi \lvert x_1 - 2 \rvert + \pi\right)\right)^2 - 1 \le x_2 \le 1 \\ 1 - \sqrt{\lvert x_1 - 2 \rvert} + 2\left(x_2 - 2 - \sin\left(6\pi \lvert x_1 - 2 \rvert + \pi\right)\right)^2 1 < x_2 \le 3 \end{cases} \end{cases}$ | $x_1 \in [-1,3], x_2 \in [1,3]$ |
| MMF6 | $\begin{cases} f_1 = \lvert x_1 - 1 \rvert \\ f_2 = \begin{cases} 1 - \sqrt{\lvert x_1 - 2 \rvert} + 2\left(x_2 - \sin\left(6\pi \lvert x_1 - 2 \rvert + \pi\right)\right)^2 - 1 \le x_2 \le 1 \\ 1 - \sqrt{\lvert x_1 - 2 \rvert} + 2\left(x_2 - 1 - \sin\left(6\pi \lvert x_1 - 2 \rvert + \pi\right)\right)^2 1 < x_2 \le 3 \end{cases} \end{cases}$ | $x_1 \in [-1,3], x_2 \in [1,2]$ |
| MMF 7 | $\begin{cases} f_1 = \lvert x_1 - 2 \rvert \\ f_2 = \begin{cases} 1 - \sqrt{\lvert x_1 - 2 \rvert} + \{x_2 - [0.3\lvert x_1 - 2 \rvert^2 \cdot \cos\left(24\pi \lvert x_1 - 2 \rvert \right. \\ \left. + 4\pi\right) + 0.6\lvert x_1 - 2 \rvert] \cdot \sin\left(6\pi \lvert x_1 - 2 \rvert + \pi\right)\}^2 \end{cases} \end{cases}$ | $x_1 \in [1,3], x_2 \in [-1,1]$ |

**Appendix C**

MOANA
ZDT1
ZDT2
ZDT3

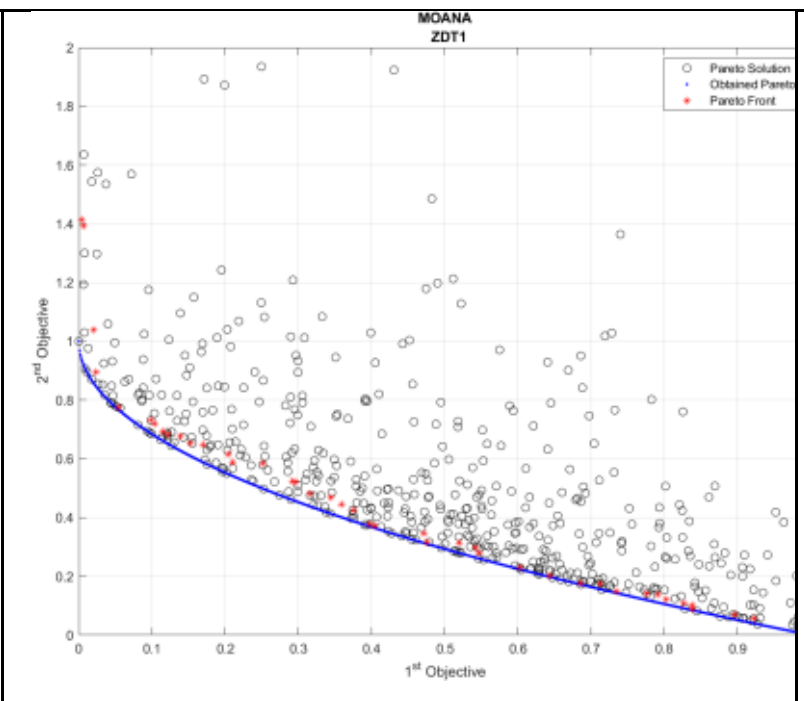

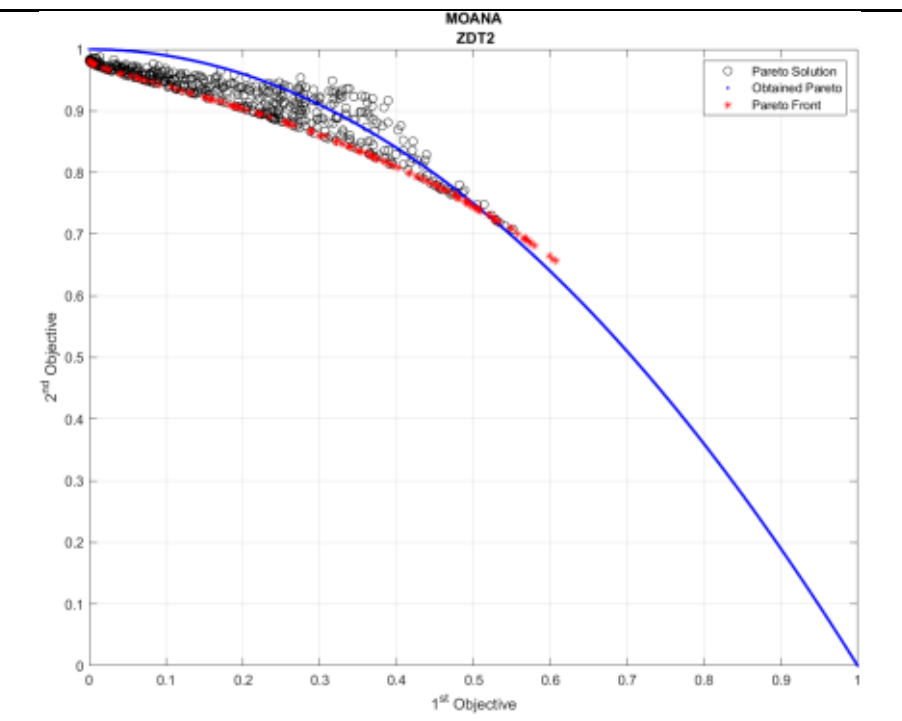

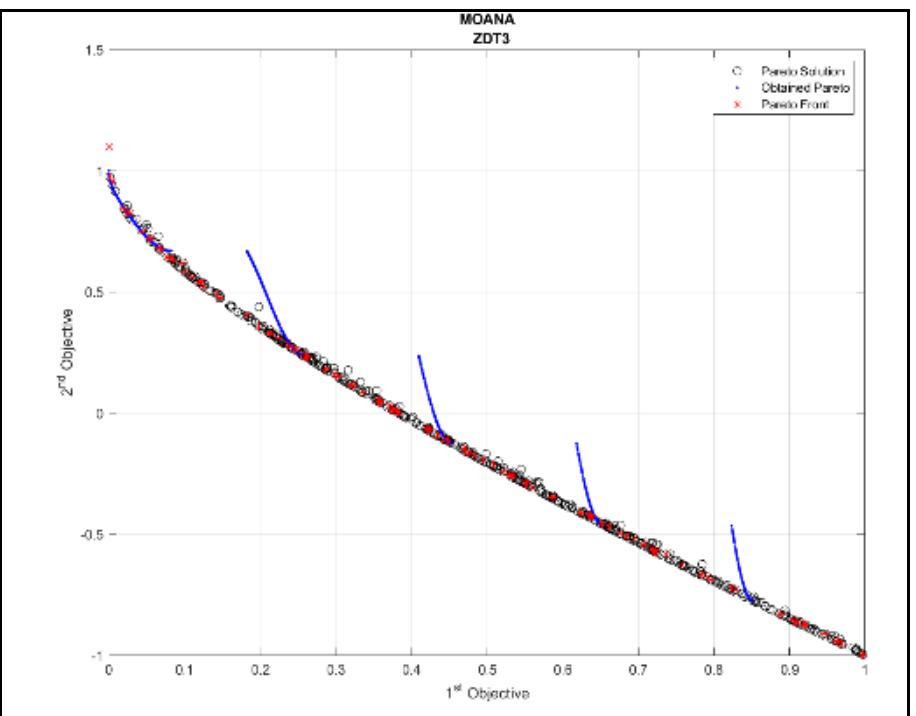

MOPSO
ZDT1
ZDT2
ZDT3

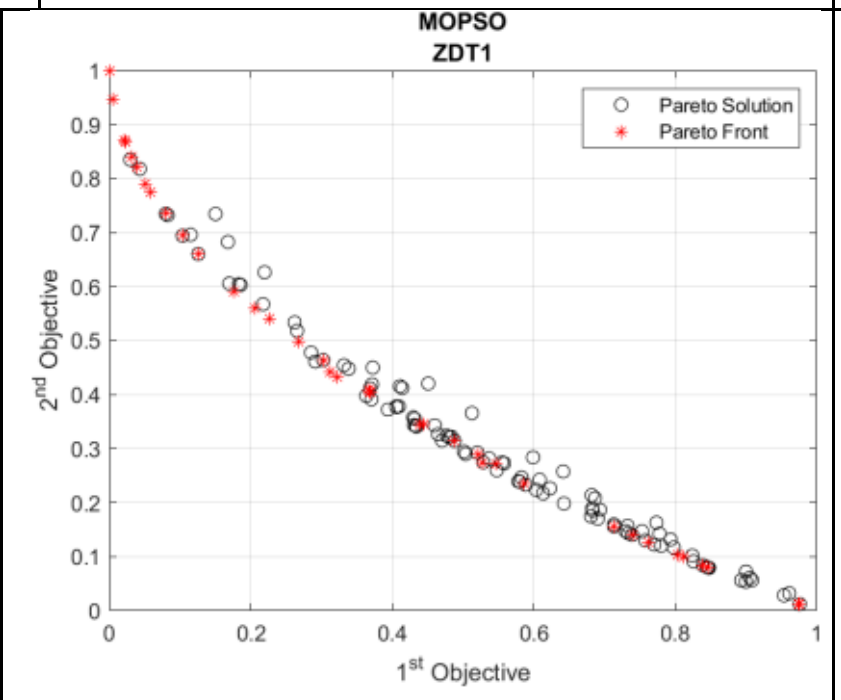

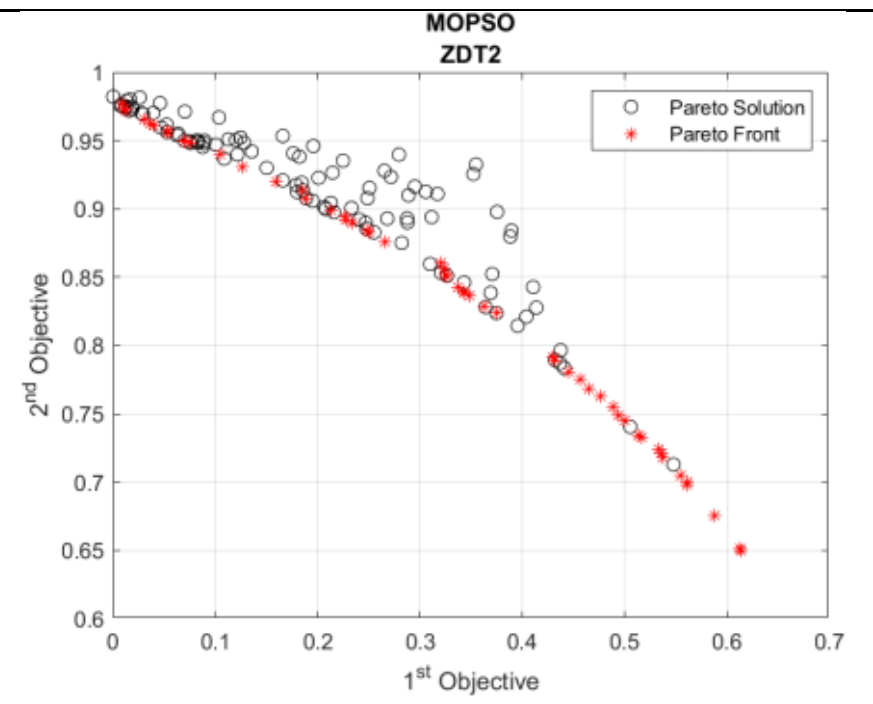

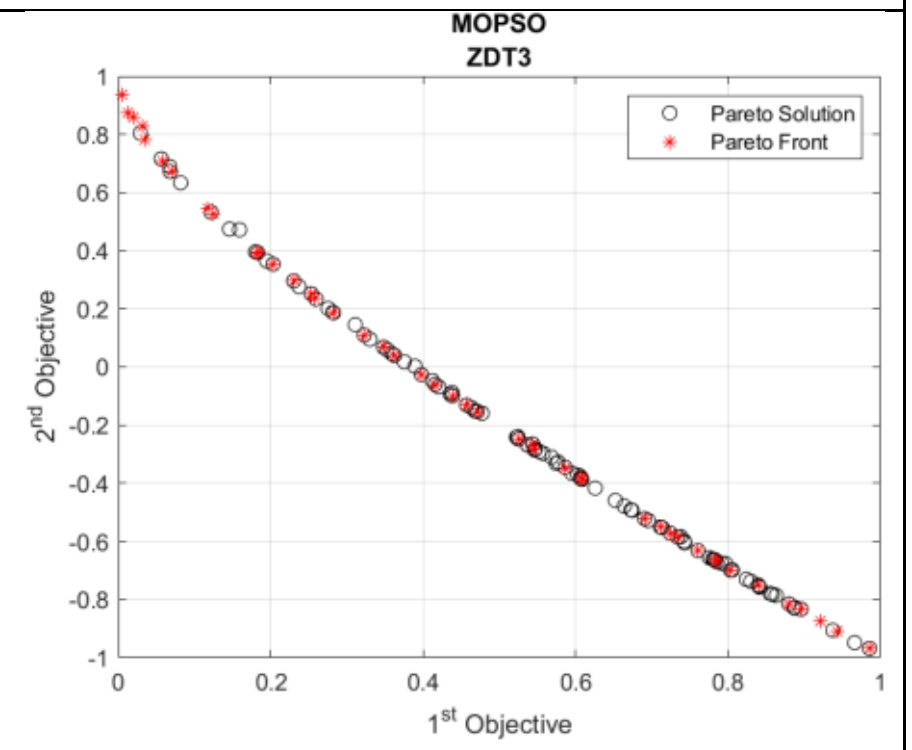

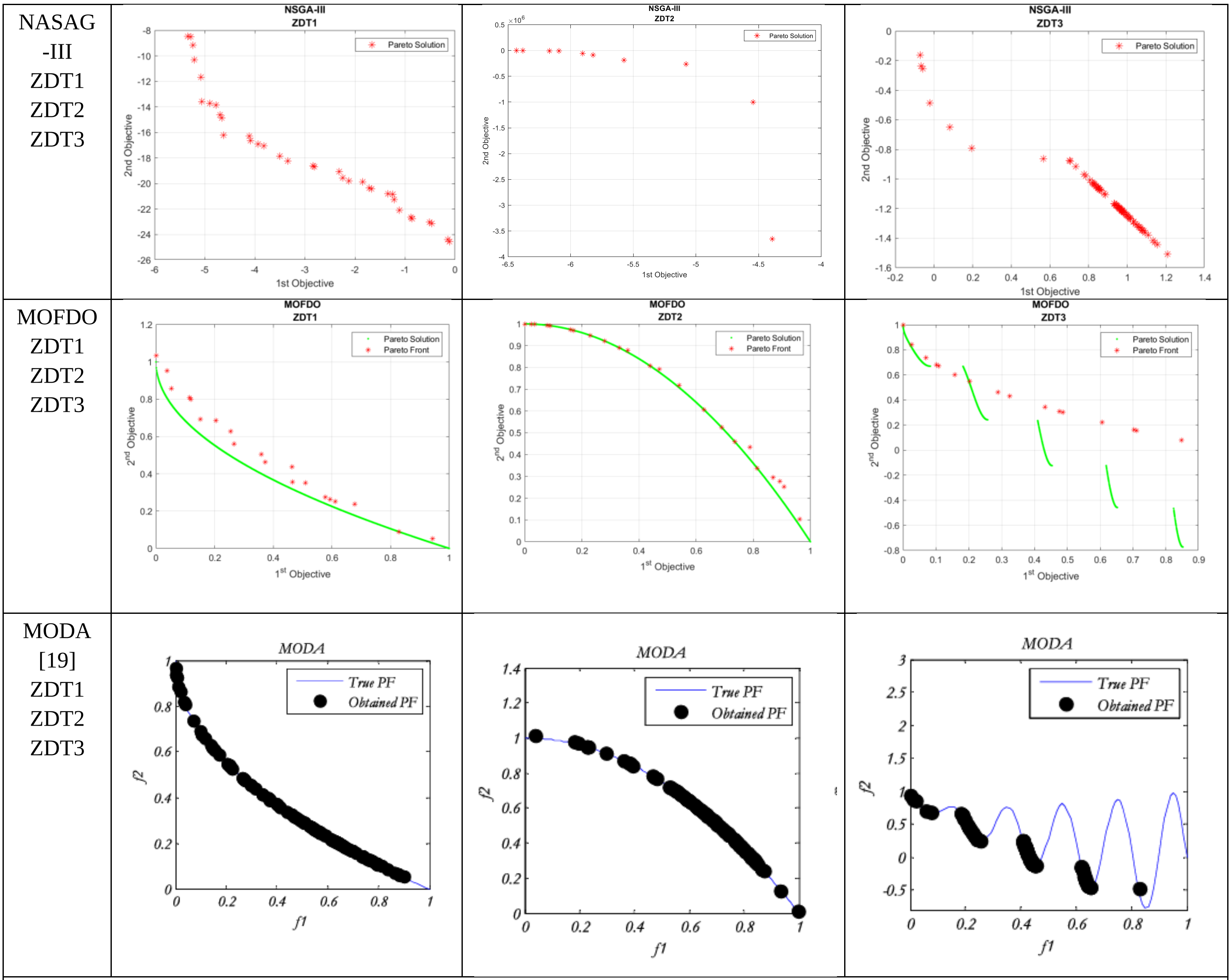

**Figure 8.** The multi-objective algorithms achieved the best Pareto optimal front across several ZDT test functions.

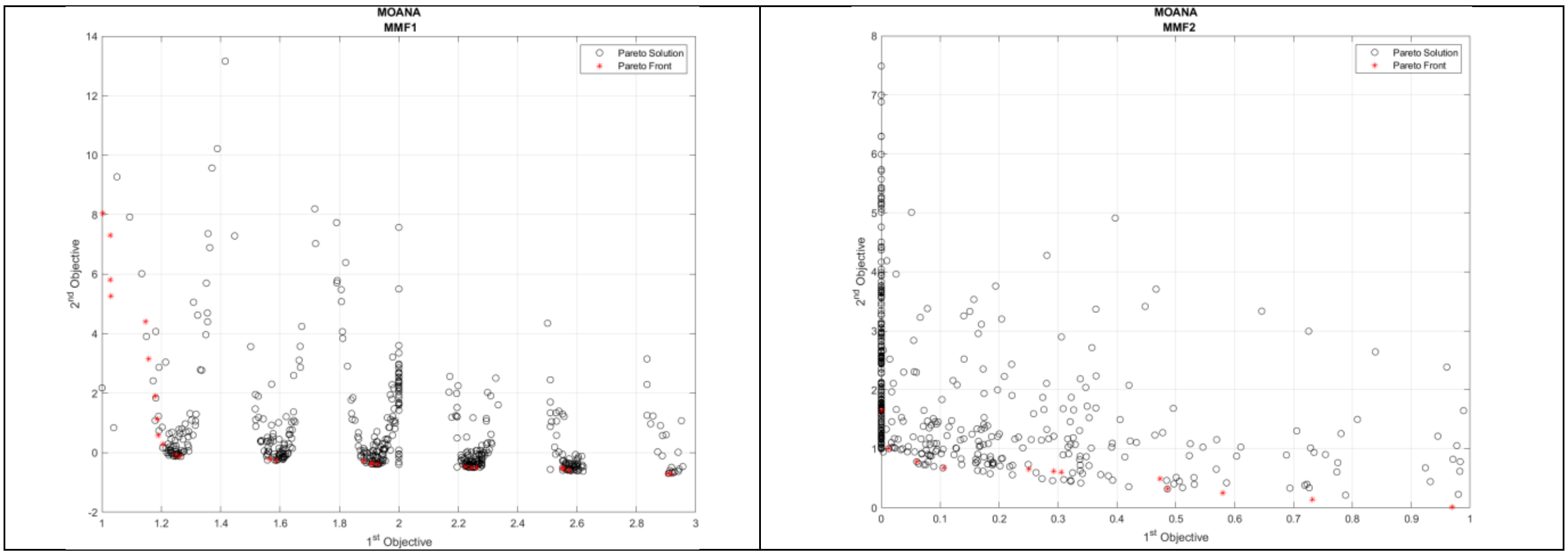

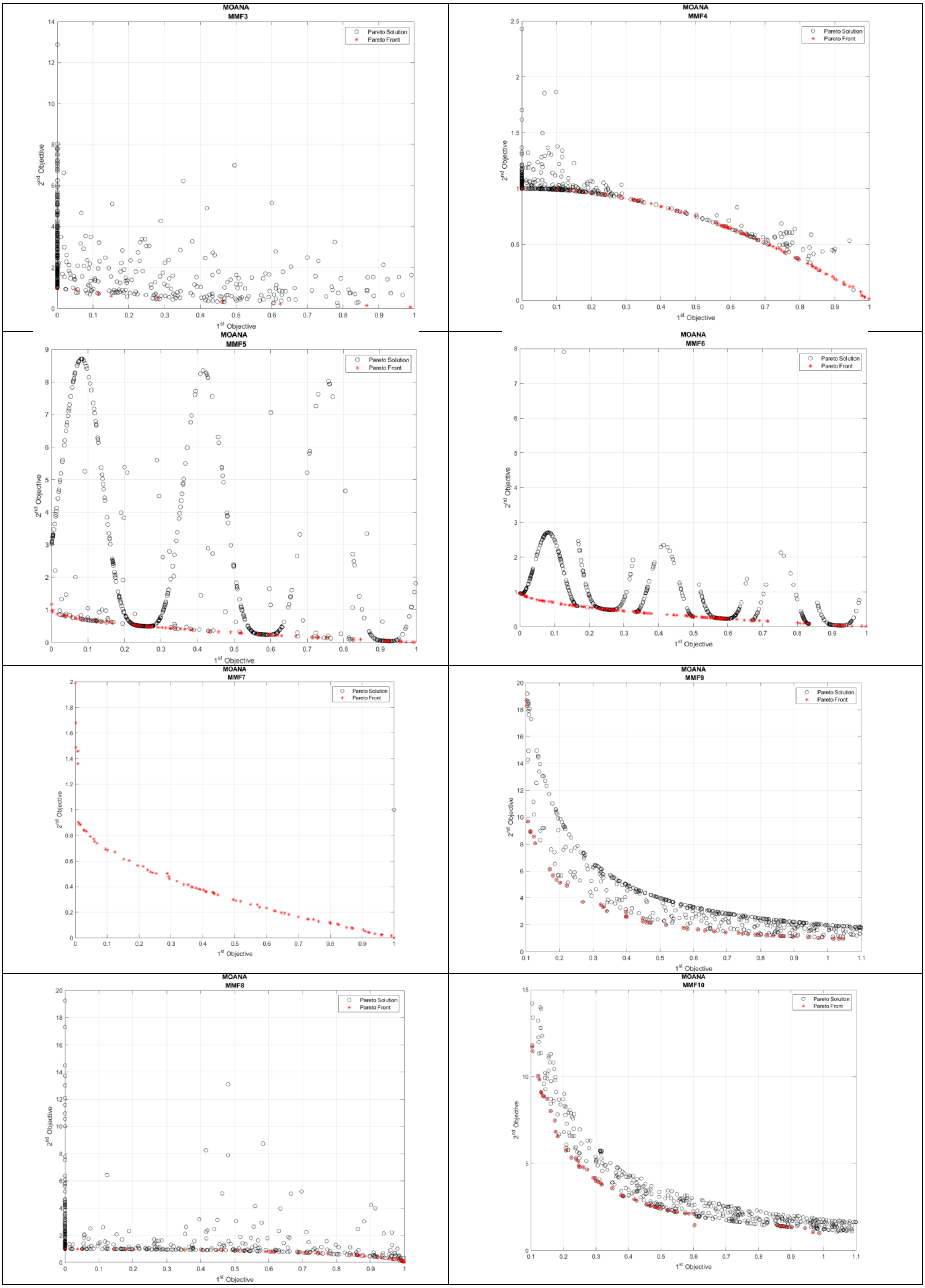

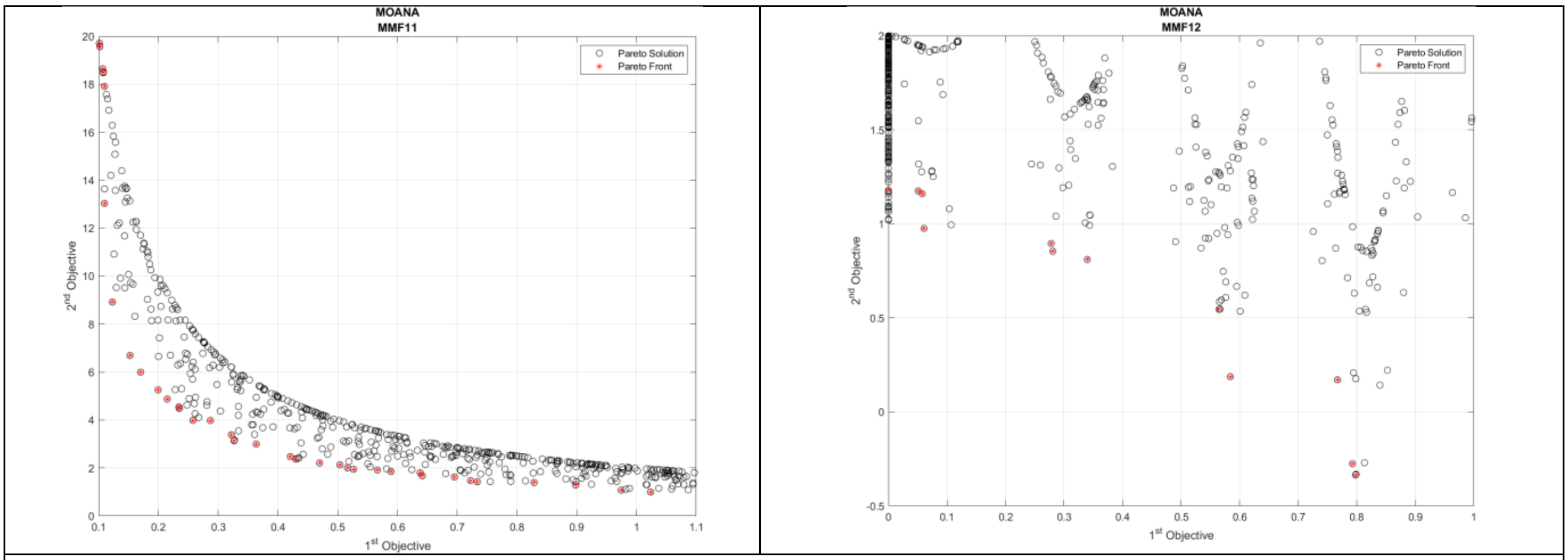

**Figure 9.**The preformance of MOANA achieved the best Pareto optimal across several MMF CEC2019 test functions.

MOANA
MMF1
MMF2
MMF4
MMF5
MMF6
MMF7
MMF8

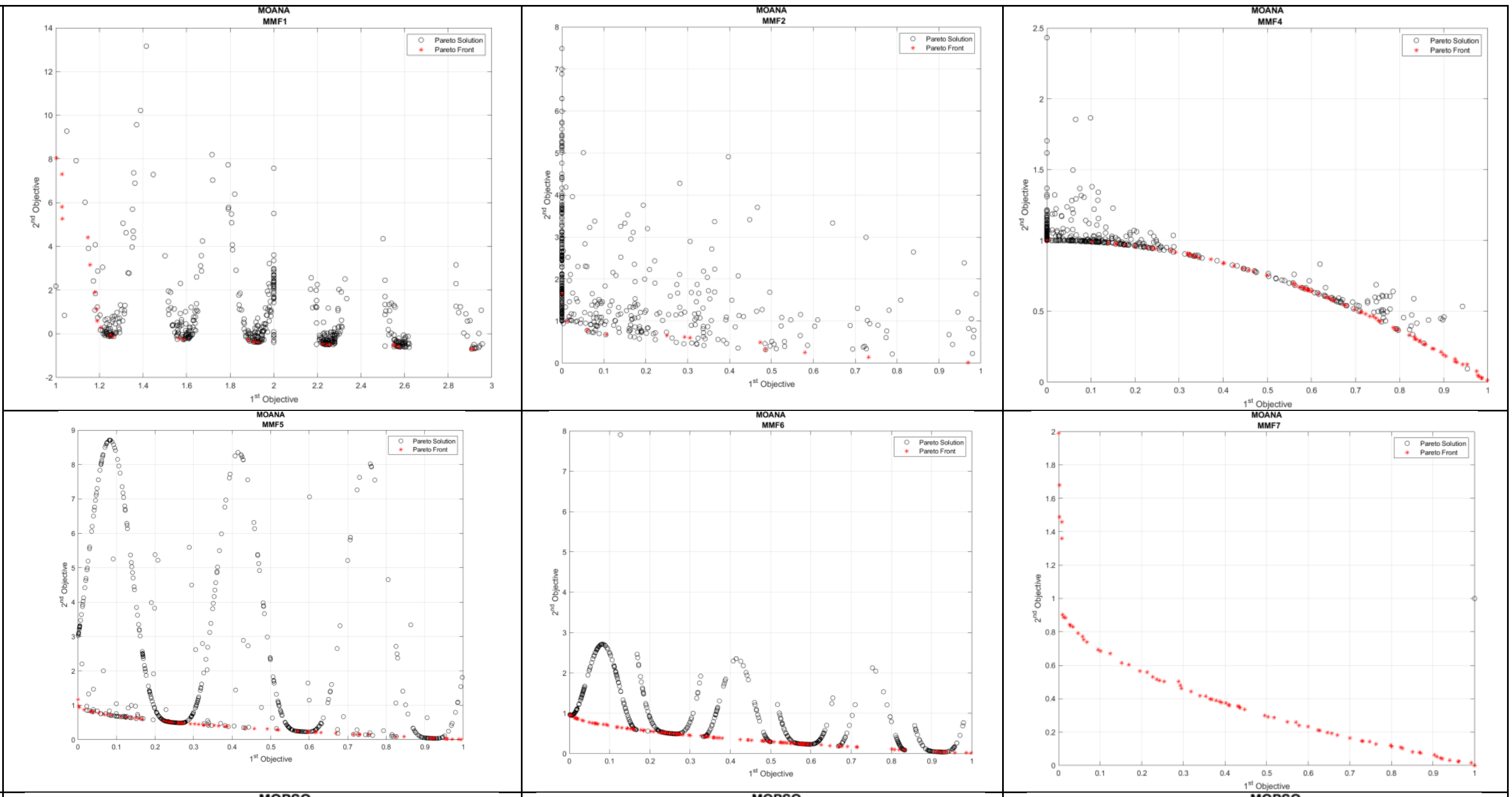

MOPSO
MMF1
MMF2
MMF4
MMF5
MMF6
MMF7

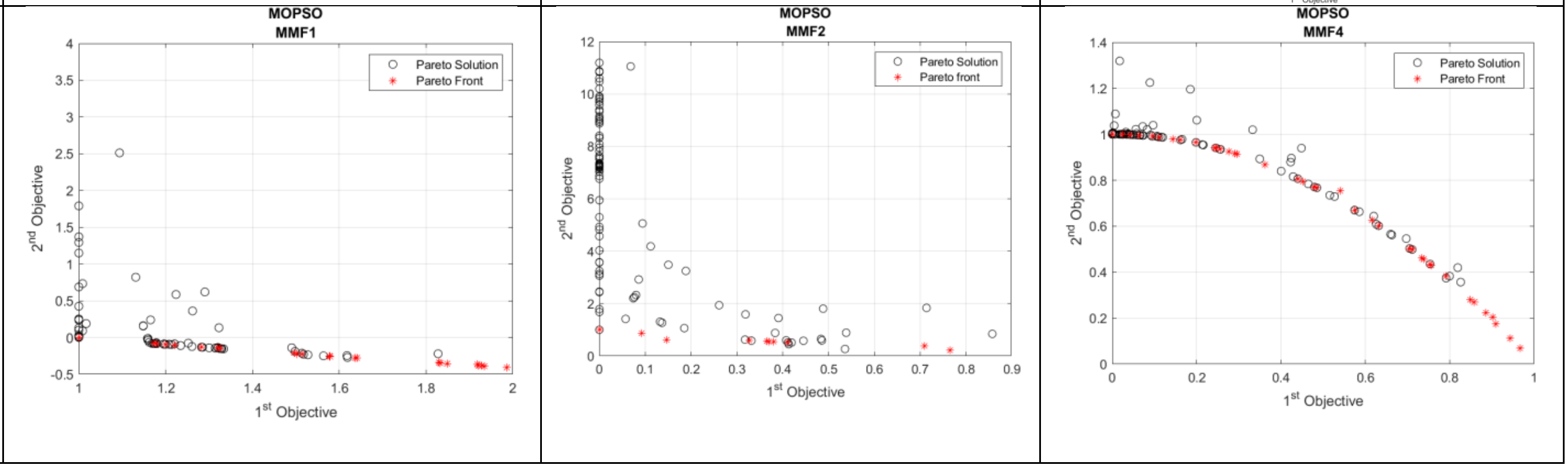

| | |
|---|---|
| MMF8 | 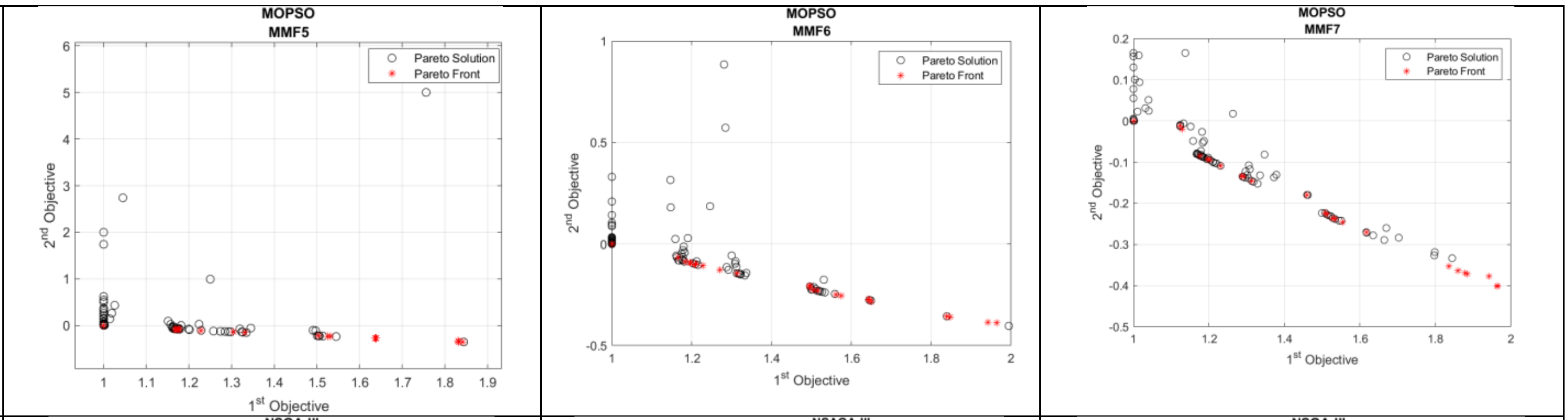  |
| NASAG-III MMF1 MMF2 MMF4 MMF5 MMF6 MMF7 MMF8 | 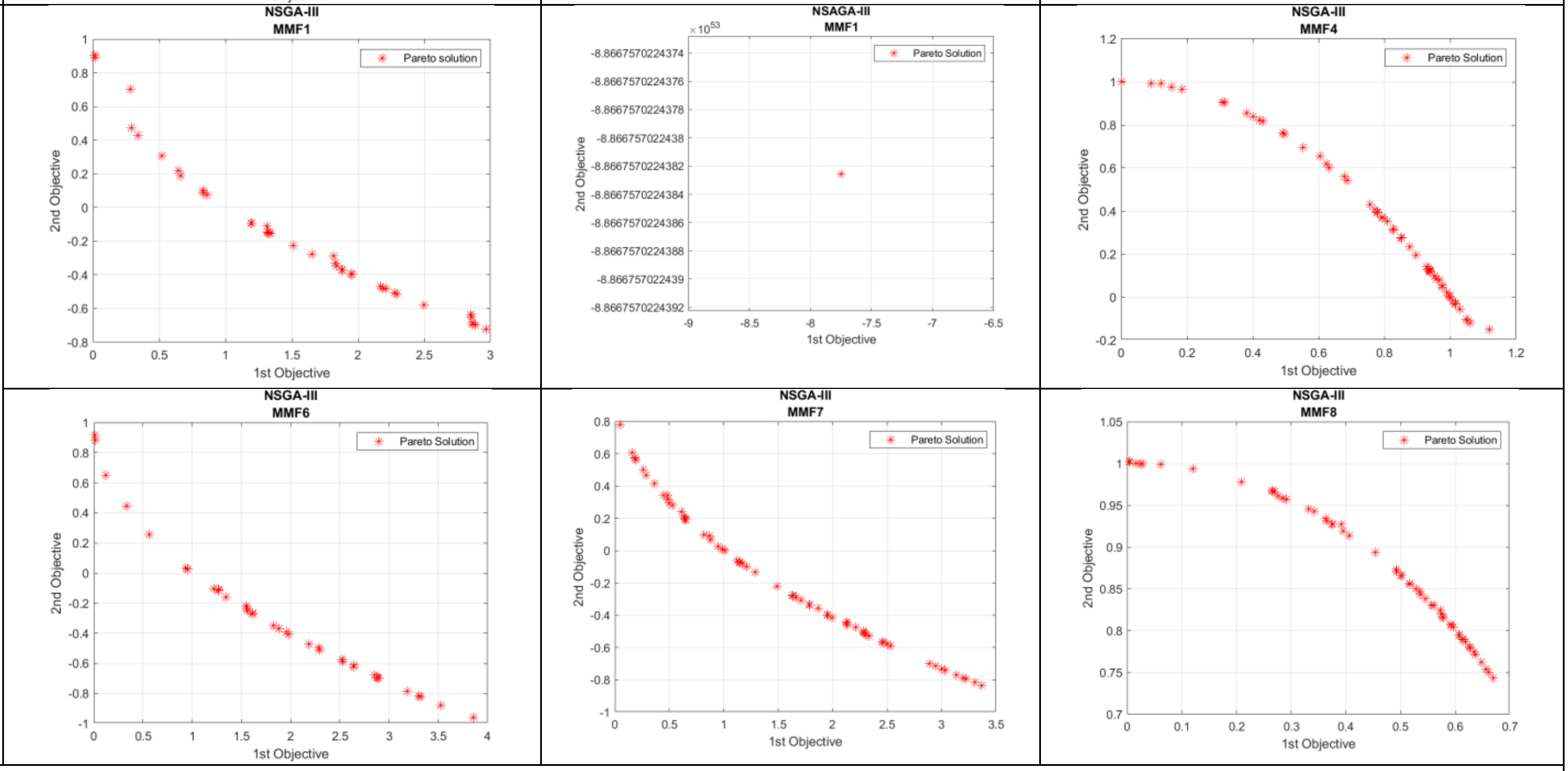  |

**Figure 10.** The multi-objective algorithms achieved the best Pareto optimal front across several MMF CEC2019 test functions.